\documentclass[10pt,journal,compsoc]{IEEEtran}
\usepackage{xr}
\usepackage{epsfig}
\usepackage{graphicx}
\usepackage{amsmath}
\usepackage{amssymb}
\usepackage{tabularx}
\usepackage{subcaption}
\usepackage{xcolor}
\usepackage{threeparttable}
\usepackage{algorithm}
\usepackage{algorithmic}
\usepackage{enumitem}
\usepackage{ragged2e}
\usepackage{multirow}
\usepackage{lipsum}
\usepackage{subcaption}
\usepackage{booktabs} 
\usepackage{color}
\usepackage{array}
\usepackage{bbm}
\usepackage{calc}
\usepackage{xspace}
\usepackage{caption}
\usepackage{colortbl}
\usepackage{hhline}
\usepackage[pagebackref=false,breaklinks=true,colorlinks,bookmarks=false]{hyperref}
\usepackage[utf8]{inputenc}
\usepackage{stfloats}

\ifCLASSOPTIONcompsoc
  \usepackage[nocompress]{cite}
\else
  \usepackage{cite}
\fi

\ifCLASSINFOpdf
\else
\fi

\usepackage{xspace}
\makeatletter
\DeclareRobustCommand\onedot{\futurelet\@let@token\@onedot}
\def\@onedot{\ifx\@let@token.\else.\null\fi\xspace}

\def\eg{\emph{e.g}\onedot} 
\def\ie{\emph{i.e}\onedot} 
 
\def\etc{\emph{etc}\onedot}

\def\etal{\emph{et al}\onedot}
\newcolumntype{P}[1]{>{\centering\arraybackslash}p{#1}}
\newcommand{\blue}[1]{\textcolor{black}{#1}}

\begin{document}
%
\title{Robust Asymmetric Heterogeneous Federated Learning with Corrupted Clients}
%
%

\author{Xiuwen Fang, Mang Ye,~\IEEEmembership{Senior Member,~IEEE}, Bo Du,~\IEEEmembership{Senior Member,~IEEE}
\IEEEcompsocitemizethanks{
\IEEEcompsocthanksitem This work is supported by National Key Research and Development Program of China (2023YFC2705700),
the National Natural Science Foundation of China under Grants (62361166629, 62176188, 62225113). (\textit{Corresponding
author: Mang Ye}.)
\IEEEcompsocthanksitem X. Fang, M. Ye, B. Du are with the School of Computer Science, Taikang Center for Life and Medical Sciences, Wuhan University, Wuhan, China, 430072. E-mail:\{fangxiuwen, yemang, dubo\}@whu.edu.cn
\IEEEcompsocthanksitem A preliminary version of this work has appeared in ICCV 2023~\cite{iccv2023aughfl}.
}
}

%
%

\markboth{IEEE Transactions on Pattern Analysis and Machine Intelligence}%
{Shell \MakeLowercase{\textit{et al.}}: Bare Demo of IEEEtran.cls for Computer Society Journals}

\IEEEtitleabstractindextext{%

\begin{abstract}
\justifying
This paper studies a challenging robust federated learning task with model heterogeneous and data corrupted clients, where the clients have different local model structures. Data corruption is unavoidable due to factors such as random noise, compression artifacts, or environmental conditions in real-world deployment, drastically crippling the entire federated system. To address these issues, this paper introduces a novel Robust Asymmetric Heterogeneous Federated Learning (RAHFL) framework. We propose a Diversity-enhanced supervised Contrastive Learning technique to enhance the resilience and adaptability of local models on various data corruption patterns. Its basic idea is to utilize complex augmented samples obtained by the mixed-data augmentation strategy for supervised contrastive learning, thereby enhancing the ability of the model to learn robust and diverse feature representations.
Furthermore, we design an Asymmetric Heterogeneous Federated Learning strategy to resist corrupt feedback from external clients. The strategy allows clients to perform selective one-way learning during collaborative learning phase, enabling clients to refrain from incorporating lower-quality information from less robust or underperforming collaborators. Extensive experimental results demonstrate the effectiveness and robustness of our approach in diverse, challenging federated learning environments. Our code and models are public available at \url{https://github.com/FangXiuwen/RAHFL}.
\end{abstract}

}

\maketitle

\IEEEdisplaynontitleabstractindextext

\IEEEpeerreviewmaketitle

\IEEEraisesectionheading{\section{Introduction}\label{sec:intro}}

\IEEEPARstart{T}{he} success of deep learning relies on large amounts of available data. In parallel, modern society contains a multitude of edge devices, such as smartphones, mobile networks, IoT devices, \etc., which can be regarded as local clients with limited private data. While collecting this data for model training can improve model performance, it risks violating privacy and security regulations.
To address this, Federated Learning (FL) was proposed~\cite{aistats2017fedavg}, which is a distributed machine learning paradigm with secure encryption technology. It enables multiple parties to collaboratively train models without sharing private data.  Recent studies have shown that FL can effectively improve model performance while preserving data privacy.
Until now, FL has been widely explored in academic research~\cite{csur2023hflsurvey, tpami2023lga} and has been successfully applied in various industrial fields, such as medical diagnosis~\cite{nature2021exam,nature2021primia,nature2021famhe,cvpr2023fedpr}, financial insurance~\cite{nature2021adversarial, intsys2021secureboost}, recommended system~\cite{sigkdd2020fedfast, flpi2020fedrecsys}, data security~\cite{nature2021adversarial}, smart city~\cite{tits2022fedstn, infsci2023flsmartcity}, \etc.

\begin{figure}[t]
\centering
   \includegraphics[width=7.5cm]{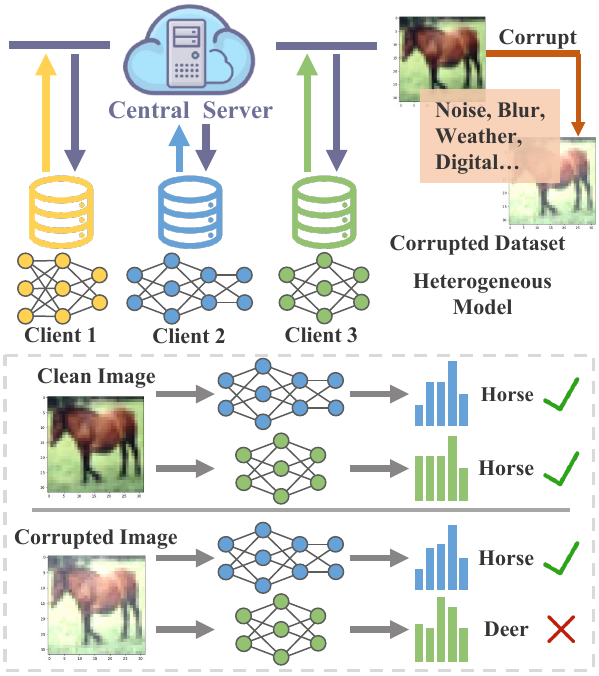}
   \caption{\small{Illustration of heterogeneous FL with data corruption, where the clients may possess different model structures and corrupted private datasets.}}
   \vspace{-5mm}
\label{fig:intro}
\end{figure}

Most existing FL algorithms assume all clients share the same local model structure, relying on weighted average model gradients or weight parameters for aggregation. For example, FedAvg~\cite{aistats2017fedavg}, FedProx~\cite{mlsys2020fedprox}, and Per-FedAvg~\cite{nips2020perfedavg} enable communication between clients by taking the weighted average of the local model parameters. However, real-world applications frequently involve devices with diverse requirements and computational capacities, which may necessitate different model architectures~\cite{icml2022fedhenn,wu2020ojcs,iccv2023fedins}, as shown in Figure~\ref{fig:intro}. Considering privacy protection, business confidentiality, etc., the clients are usually reluctant to disclose or share model design details with others. In the model heterogeneous setting, conventional aggregation methods are limited, resulting in difficulties in model communication, which poses a huge challenge to maintaining the efficiency and effectiveness of the collaborative learning process. Existing works exhibit different methodologies in handling communication among heterogeneous clients.
FedMD~\cite{nips2019fedmd} facilitates model communication based on the average output class scores of the local models. 
FedDF~\cite{nips2020feddf} capitalizes on unlabeled or synthetically generated data to distill knowledge across clients. 
RHFL~\cite{cvpr2022rhfl} transfers knowledge by aligning feedback outputs of local models on public irrelevant data. 
HeteroFL~\cite{iclr2021heterofl} dynamically distributes a subset of the global model parameters to local models, optimizing for the different computational and communication capacities of the participating clients.
KT-pFL~\cite{nips2022ktpfl} implements personalized knowledge transfer by employing a parameterized knowledge coefficient matrix.
FedProto~\cite{aaai2022fedproto} utilizes class prototypes~\cite{tpami2023sketchtrans} to facilitate the aggregation and communication processes in FL. 
Nonetheless, the success of these strategies presupposes the availability of clean, uncorrupted images in all private datasets, which is rarely satisfied in practical scenarios.

In real-world FL systems, it is costly and unrealistic to require all clients to have high-quality clean data. In the practical process of data collection, transmission or storage, many factors may cause data corruption, \eg, random noise, compression artifacts, environmental conditions and other unexpected distortions. Consequently, there is inevitably data corruption on the client side, which introduces errors and biases in the local training, causing models to learn the wrong patterns and degrading overall performance. In addition, there may be free-rider participants~\cite{lin2019freerider, tpami2024flsurvey} in the FL system, whose existence will exacerbate the potential for data corruption. Free-riding participants refer to clients that do not make effective contributions to the federated collaborative training but still benefit from the collective results. This behavior poses a significant challenge to data privacy protection and user fairness in FL. Therefore, the honest clients may deliberately provide low-quality data to protect their privacy and interests. Due to the existence of data corruption, clients will iteratively learn and share wrong knowledge, causing models to update in the wrong direction, which will significantly degrade the performance of the FL system and hinder the actual deployment of the model. Therefore, developing robust FL algorithms is essential for the successful implementation of FL.
In single-model machine learning, several approaches have been developed to mitigate the impact of data corruption, which can generally be divided into four categories: data augmentation~\cite{wacv2022agmax,lopes2019gausaug,nips2021monkey}, noise injection~\cite{nips2021nrnn,liu2019neuralsde,cvpr2020neuralode}, pre-training~\cite{icml2019ptt,icip2022bandlimit,icml2020unilmv2}, and frequency regularization~\cite{iccv2021rohl,ijcai2022jafr}. However, the multi-model scenario of FL is more complex, making it more challenging to overcome data corruption. We expect to fully learn local knowledge and resist multiple corruptions within the FL framework. Therefore, \textit{how to mitigate model performance degradation from data corruption during the local learning phase} is an important challenge.

\begin{figure}[t]
\centering
   \includegraphics[width=8.3cm]{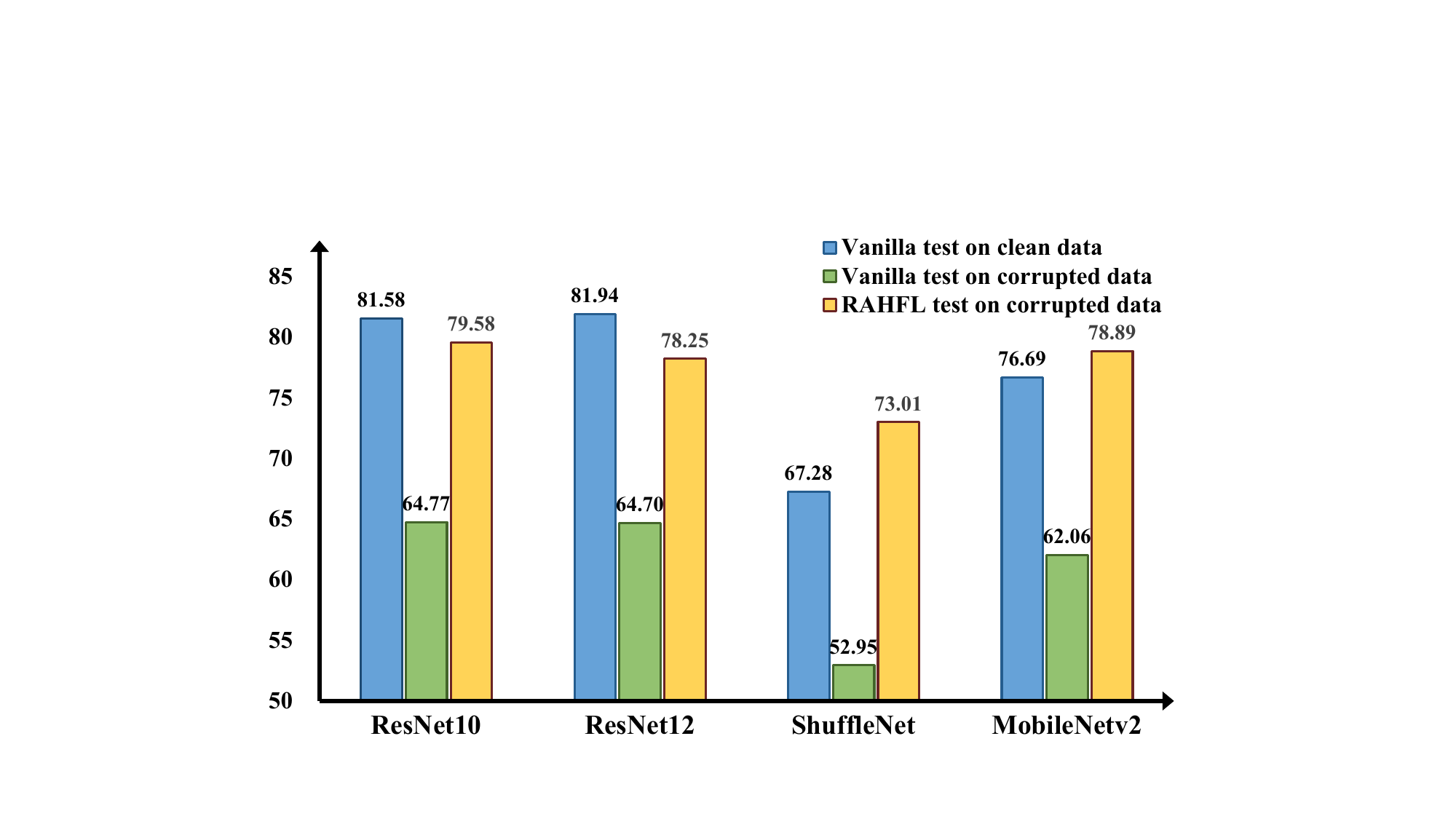}
   \caption{\small{Model performance on clean vs. corrupted test data. We observe that corrupted data is more susceptible to misprediction than clean data. RAHFL outperforms vanilla models on corrupted data while maintaining accuracy on clean data.}}
   \vspace{-3mm}
\label{fig:prove}
\end{figure}

In FL scenarios, variable data quality and model performance among clients lead to unbalanced learning. High-quality clients can be negatively affected by low-quality clients. The errors and biases introduced by corrupted data propagate and amplify during the collaboration, thereby reducing the overall performance of the FL system. Furthermore, in heterogeneous FL scenarios, the local models of clients may have different model architectures, and heterogeneous models have individual decision boundaries. The corrupted samples are usually more difficult to classify correctly than clean samples, as shown in Figure~\ref{fig:prove}. Therefore, for the same corrupted image, the predictions of different models may vary significantly (Figure~\ref{fig:intro}), resulting in inconsistent optimization directions of the models. In the collaborative communication of heterogeneous models, clients may inadvertently reinforce incorrect predictions from unreliable models.
Therefore, data corruption in model heterogeneous scenarios encounters an additional problem, \ie, \textit{how to minimize the corrupted knowledge learned from unreliable clients while achieving heterogeneous models communication. }

This paper introduces a novel framework, termed Robust Asymmetric Heterogeneous Federated Learning (RAHFL), to address the above-mentioned challenges. RAHFL consists of two stages: 1) To mitigate the model performance degradation caused by local data corruption, we adopt a random mixed data augmentation strategy with consistency loss, and then design a Diversity-enhanced supervised Contrastive Learning (DCL) method. The method utilizes complex augmented samples obtained by random mixed data augmentation for supervised contrastive learning, thereby promoting local models to learn more diverse feature representations. To prevent potential failure due to excessive distortion of complex augmented images, we maintain consistency in the similarity probability distribution between simple and complex augmentation patterns. By learning diverse complex image patterns, the resilience and adaptability of local models to various forms of corrupted data are enhanced. 2) To prevent the global paralysis caused by learning corrupted knowledge from others, we design a general Asymmetric Heterogeneous Federated Learning (AsymHFL) method, that enables robust heterogeneous model communication. The method does not force participating clients to learn in pairs, but allows selective one-way learning to avoid clients incorporating low-quality information from poorly performing collaborators. The main contributions in this work are as follows:

\begin{itemize}
\item We introduce RAHFL to address the challenging robust heterogeneous FL problem. It can effectively handles both local and external data corruption while achieving robust communication among heterogeneous models.
\item We propose a DCL strategy, which effectively uses complex enhanced samples for supervised contrastive learning to prevent model performance degradation caused by local data corruption.
\item We design a general method, AsymHFL, which enables clients to asymmetrically transfer knowledge to avoid learning low-quality information from others, achieving robust heterogeneous model communication.
\item Extensive experiments are conducted to demonstrate the effectiveness and robustness of our framework in diverse and challenging FL environments. The framework consistently outperforms State-Of-The-Art (SOTA) methods.
\end{itemize}

A preliminary conference paper version~\cite{iccv2023aughfl} has been published in ICCV 2023. In this journal version, we provide the following three major improvements: First, we introduce a diversity-enhanced supervised contrastive learning strategy to achieve robust feature representation learning for local models. This strategy fully captures the diverse patterns contained in complex augmented data generated by random mixed augmentation, thereby enhancing the robustness of models to unseen data corruption. Second, we introduce the AsymHFL strategy, which enables clients to selectively integrate knowledge only from higher-performing models during the collaborative update phase. This approach prevents the propagation of low-quality information, thereby maintaining overall learning efficiency and accuracy. Finally, we further demonstrate the robustness and applicability of RAHFL in complex FL scenarios through extensive experiments. 
We conduct extensive experiments and comprehensive analysis involving both heterogeneous and homogeneous models, as well as scenarios with heterogeneous data, providing detailed comparisons with SOTA methods. 
We introduce additional experiments with numerous clients to demonstrate the scalability and effectiveness of the proposed method in complex FL environments.

\section{Related Work}
\textbf{Federated Learning.} FL is a decentralized learning paradigm first introduced in 2017 by McMahan \etal ~\cite{aistats2017fedavg}, which allows multiple clients to collaboratively train local models without compromising data privacy. In the initial FL algorithm, FedAvg, clients independently train their local models with private datasets and share parameters with the central server. Then, the server averages these parameters of all clients and this iterative process continues for several rounds. FedProx~\cite{mlsys2020fedprox} lightens the computational burden of clients by adjusting the local training epochs, enhancing training stability and accelerating convergence. To handle the problem that excessive information transfer weakens the system stability, Wu \etal~\cite{aaai2022smartidx} propose a novel federated compression algorithm SmartIdx. It develops a kernel-based parameter selection strategy and a parameter compression algorithm, reducing the communication overhead and boosting communication efficiency. FedRCL~\cite{cvpr2024fedrcl} mitigates local update bias by applying supervised contrastive learning to FL, penalizing overly similar sample pairs to prevent representation collapse.

Unlike traditional homogeneous FL, several clients expect to design local models themselves, which brings about model heterogeneous FL. Most current approaches for heterogeneity~\cite{arxiv2020cfd, arxiv2019cornus, pmlr2022fedhenn, nips2020fedgkt} leverage knowledge distillation technique~\cite{nips2015kd} to facilitate communication between clients without direct data or parameter sharing. FedMD~\cite{nips2019fedmd} enables heterogeneous models to interact by learning the average aggregated soft labels across all clients. Several methods~\cite{arxiv2019cornus,nips2022ktpfl} utilize public data for knowledge distillation, while others like FedDF~\cite{nips2020feddf}, FCCL~\cite{cvpr2022fccl} and RHFL~\cite{cvpr2022rhfl} leverage unlabeled data to implement knowledge distillation and transfer among clients. HeteroFL~\cite{iclr2021heterofl} assigns personalized subsets of global model parameters to clients based on their computational and communication capabilities, rather than allowing clients to independently design models. Several approaches~\cite{aaai2022fedproto,cvpr2024fedktl} exploit prototype learning, aggregate local prototypes of clients to obtain a global prototype, and facilitate federated learning by constraining the distance between local and global prototypes.

In summary, SOTA model heterogeneous FL methods are generally developed under the assumption that the training images are flawless, overlooking the issue of data corruption. This oversight can hinder model convergence and degrade performance when corrupted data is present. Given the lack of research addressing data corruption in heterogeneous FL, our objective is to develop an algorithm capable of handling corrupted data on client devices.

\noindent\textbf{Data Corruption Learning.} Data corruption is a common issue in real-world scenarios~\cite{eccv2020ant, iclr2021edgeaug}, occurring during data collection, storage, and transmission. It can also result from free-riders in FL. Dodge \etal~\cite{qomex2016sensitivity} point out that models are highly sensitive to corrupted datasets, which significantly degrades their performance. SOTA methods employ several techniques to mitigate these adverse effects:

1) \textit{Data augmentation.} These methods aim to enhance the robustness and reduce the generalization error by augmenting corrupted data. MixUp~\cite{iclr2018mixup} realizes data augmentation by a linear combination of any two training data. Verma \etal~\cite{icml2019manifoldmixup} extend the linear combination algorithm of MixUp with Manifold Mixup, which smooths the decision boundaries of models. Teach Augment~\cite{cvpr2022teachaugment} improves the generalizability of models by using an adversarial model to prevent feature loss caused by excessive data augmentation. Besides, AugMax~\cite{nips2021augmax} learns adversarial mixtures of random augmentations to generate harder augmented samples.
2) \textit{Noise injection.}~\cite{uai2022effectiveness} The technique can be considered as a form of data augmentation~\cite{erichson2022noisymix}, which prevents the model from overfitting to corrupted data by adding noise to the training process. Early algorithms~\cite{nc1996datanoise} add noise directly to training data to improve robustness against data corruption. In subsequent works, noise is added to other parts of the models. Gulcehre \etal~\cite{icml2016noisyactivation} inject noise to activation functions which effectively accelerates the convergence of models trained on corrupted data. Some methods~\cite{nips2020ergnj,iclr2021nfm} inject noise to fully connected layers of MLP and CNN to improve model robustness against data corruption. 
3) \textit{Pre-training.} Training models on diverse datasets with large domain gaps also enhance robustness. Hendrycks \etal~\cite{icml2019ptt} comprehensively prove that the pre-trained models have stronger robustness and generalization. Noisy Student Training~\cite{cvpr2020nst} incorporates a pre-training phase to boost resilience to data corruption.
4) \textit{Frequency regularization.} The approaches aim to leverage the frequency characteristics of data corruptions to enhance the robustness of the model to different frequency. RoHL~\cite{iccv2021rohl} applies regularization to minimize the total variation of the convolutional feature map to improve high-frequency robustness. Chan \etal~\cite{ijcai2022jafr} explore the impact of frequency bias on model robustness and adopt Jacobi frequency regularization to train models biased toward low-frequency or high-frequency features.

Currently, some works attempt to address data corruption in FL~\cite{tist2022arfl, stripelis2022perweight}, typically using empirical risk to set the aggregation weights of clients. However, these methods are designed for homogeneous models and cannot be directly applied to heterogeneous FL scenarios. In summary, while existing techniques focus on centralized training and homogeneous FL, there is a lack of approaches addressing data corruption in model heterogeneous FL.

\section{The Proposed Method}
\begin{figure*}[t]
\centering
   \includegraphics[width=\textwidth]{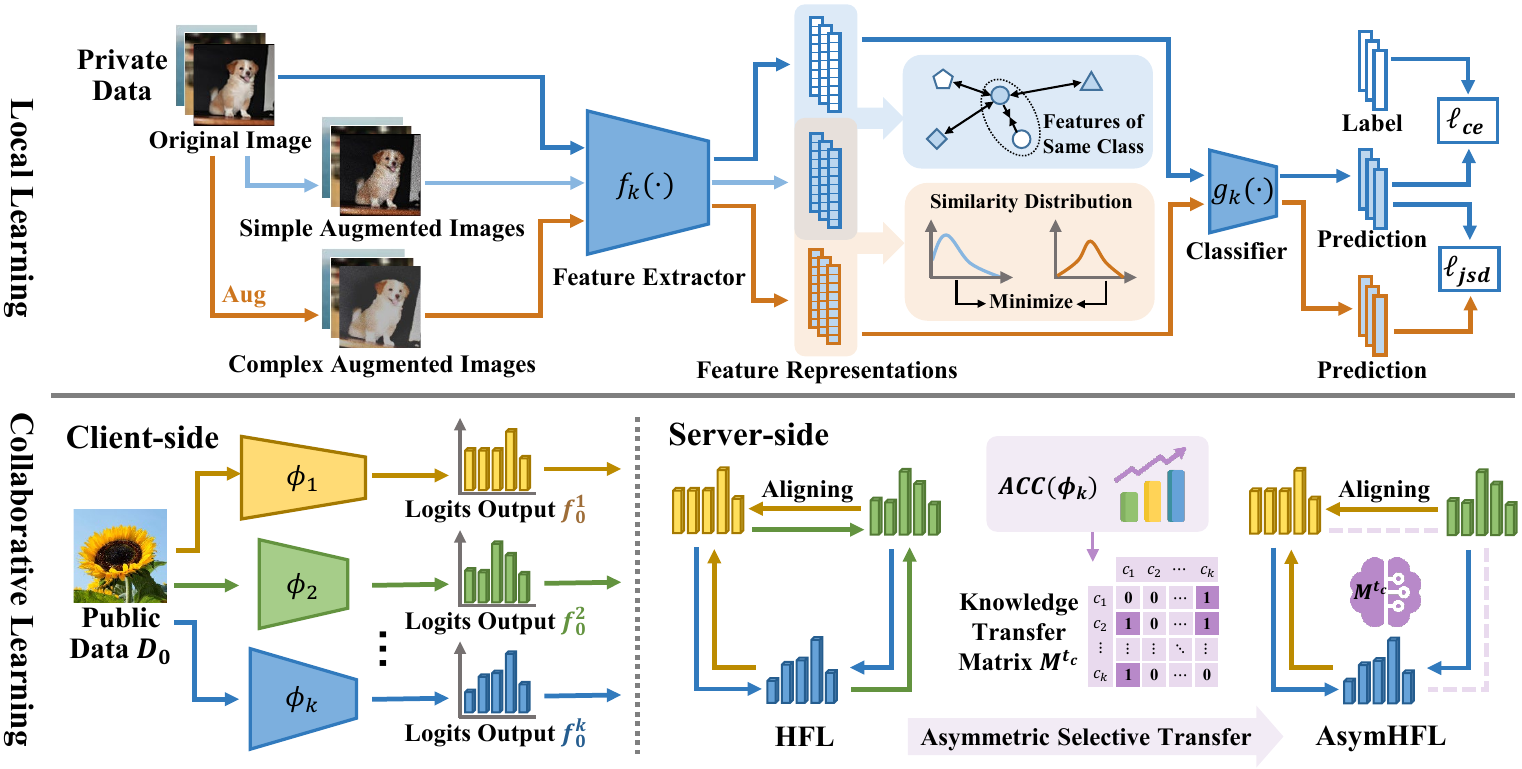}
   \caption{Overview of the RAHFL framework. The RAHFL framework consists of a local learning phase and a collaborative learning phase. In the local learning phase, random mixed augmentation and DCL are introduced to mitigate the negative impact of local data corruption (Sec.\ref{sec:localupdate}). In the collaborative learning phase, clients utilize the AsymHFL strategy for asymmetric and efficient knowledge transfer, thereby avoiding the integration of low-quality information (Sec.\ref{sec:colupdate}). The diagram shows the differences between traditional HFL and the new AsymHFL approach.}
   \vspace{-3mm}
\label{fig:method}
\end{figure*}

In this section, we elaborate on RAHFL (Figure~\ref{fig:method}), our proposed robust asymmetric heterogeneous FL method for addressing the data corruption issues in heterogeneous FL.

\subsection{Preliminaries}
In this paper, we consider the $C$-class image classification task and assume a FL system with $K$ clients and one server. The $k$-th client $c_k$ has a private dataset $D_k=\{(x_i^k,y_i^k)\}_{i=1}^{N_k}$ with $|D_k|=N_k$, where $x_i^k$ is an image, $y_i^k\in \{1,2,...,C\}$ is the corresponding label and $N_k$ refers to the number of samples in $D_k$. To protect privacy, the client $c_k$ never shares the private dataset $D_k$ with the server and other clients $c_{k'\neq k}$. In the context of model heterogeneity, each client $c_k$ has an independently designed local model $\phi_k$ with a unique structure. The model $\phi_k(\theta_k)$ with parameters $\theta_k=\{u_k,v_k\}$ contains a feature extractor $f_k(u_k)$ with parameters $u_k$ and a classifier $g_k(v_k)$ with parameters $v_k$.

In FL scenarios with data corruption, the private datasets of clients inevitably contain some samples with different corruption types and different severity levels. Hence, to explore the data corruption problem in heterogeneous FL, we consider a set of corruption functions $\mathcal{E}$. Each client $c_k$ has a corrupted private dataset $\tilde{D}_k=\{(\tilde{x}_i^k,y_i^k)\}_{i=1}^{N_k}$, where $\tilde{x}_i^k$ is an image sample that can be a clean image $x_i^k$ or a corrupted image $\tilde{x}_i^k=\varepsilon(x_i^k)$ with $\varepsilon\in\mathcal{E}$. For communication purposes, the server has an unlabeled public dataset $D_0=\{x_i^0\}_{i=1}^{N_0}$ with $|x_i^0|=N_0$, which can be directly accessed by all clients.

FL generally consists of the local learning phase and the collaborative learning phase to balance local and global knowledge. Here we denote the rounds of local update and collaborative update by $T_l$ and $T_c$, respectively. It is necessary to enhance the stability and robustness of classifiers against local corruption when using private datasets for local updates. Besides, due to the difficulty of correctly classifying corrupted data, heterogeneous models may have inconsistent prediction outputs for the same corrupted images, which can be expressed as $\phi_{k_1}(\tilde{x},\theta_{k_1})\neq \phi_{k_2}(\tilde{x},\theta_{k_2})$. Therefore, we need to prevent the client from learning corrupted predictions from others during the collaborative update phase. 

\blue{The objective is to obtain an optimal set of model parameters $\{\theta_1,\theta_2,...,\theta_K\}$ that minimizes the empirical risk $\mathbb{E}(\phi)$, as follows:}

\begin{equation}\label{eq:objective}
\mathop{\arg\min}\limits_{\theta_k} \mathcal{L}(\theta_k)=\mathbb{E}_{(\tilde{x},y)\sim D_k}[\ell(\phi_k(\tilde{x},\theta_k),y)],
\end{equation}
where $\mathbb{E}_{(\tilde{x},y)\sim D_k}$ is the empirical loss of $c_k$, and $\ell (\cdot)$ is the loss function.


\subsection{Robust Local Learning with Data Corruption}\label{sec:localupdate}

To address local data corruption in FL, we introduce novel techniques aimed at enhancing the robustness of local learning processes. First, we employed a random mixed augmentation strategy~\cite{iclr2020augmix} with consistency loss to improve the resilience of local models against varied data distortions. Then, a diversity-enhanced supervised contrastive learning approach is designed to further strengthen local model generalization by leveraging complex augmented samples, promoting robustness to unseen data corruptions.

\textbf{Random Mixed Augmentation with Consistency Loss.} 
In the local learning phase, we aim to mitigate the negative impacts of data corruption while preserving local knowledge acquisition. We leverage data augmentation~\cite{devries2017cutout,iccv2019cutmix,iclr2018mixup,cvpr2019autoaugment,cvpr2020randaugment}, a widely-used technique that enhances model generalization and robustness against corruption by generating diverse training samples~\cite{iclr2019imagenettrained}. To further strengthen the resilience of the local model against data corruption, we learn the solution strategy from Hendrycks \etal~\cite{iclr2020augmix}.

We select a set of traditional data augmentation operations $\mathcal{A}$, which includes nine operations, such as autocontrast, equalize and rotate. These operations are chosen to ensure they differ from the original corruption patterns $\mathcal{E}$ in the private datasets. The operations are applied with varying magnitudes. From this set, one to three augmentation operations $a \sim \mathcal{A}$ are randomly selected and stacked to create multiple augmentation operation sequences $Seq$ with different depths. A set of weights $(w_1,...,w_\mathcal{S})$ is randomly sampled from the $Dirichlet(\alpha,...,\alpha)$ distribution. $\mathcal{S}$ is the number of sequences. The resulting augmented images $Seq(x)$ are then mixed according to the corresponding weights $w$.
This process can be formulated as follows:
\begin{equation}\label{eq:AugMix_mix}
\begin{aligned}
&x_{seq}=\sum\nolimits_{i=1}^\mathcal{S}w_i\cdot Seq_i(x), \\
Seq&\sim\{a_1,a_1\oplus a_{2},a_1\oplus a_2\oplus a_3\}, \\
\end{aligned}
\end{equation}
where $a_1,a_2,a_3\sim\mathcal{A}$, and $\oplus$ denotes the operation combinations. This diverse combination of augmentation operations, magnitudes, sequences, and random mixing ensures a wide variety of augmented images. To preserve the original semantic information, we combine the mixed image with the original image, resulting in the final augmented image:
\begin{equation}\label{eq:AugMix_mix2}
\begin{aligned}
x'=\eta\cdot x+(1-\eta)\cdot x_{seq},
\end{aligned}
\end{equation}
where $\eta$ refers to the weight randomly sampled from the $Beta(\alpha,\alpha)$ distribution and $\alpha$ is the scaling parameter of the Dirichlet and Beta distribution, which defaults to 1. This weighting helps maintain a balance between the augmented and original images, preserving essential features while introducing variability.

To further improve model stability, we employ Jensen-Shannon Divergence (JSD) consistency loss, a symmetric and smooth variant of Kullback-Leibler (KL) divergence. The JSD loss constrains the output stability of the classifier on different augmentations of the same image. For each original image $x$, two different augmented samples $x'_1$, $x'_2$ are generated by the above method. The JSD consistency loss can be specifically expressed as:
\begin{equation}\label{eq:AugMix_jsd}
\begin{aligned}
\ell_{jsd}=\frac{1}{3}(KL(\phi||\bar{\phi})+KL(\phi'_1||\bar{\phi})+KL(\phi'_2||\bar{\phi})).
\end{aligned}
\end{equation}
Here, $\phi$, $\phi'_1$, $\phi'_2$ denote the output distribution $\phi(x,\theta)$, $\phi(x'_1,\theta)$, $\phi(x'_2,\theta)$ of the original image $x$ and two augmented images $x'_1$, $x'_2$ on the local model respectively, and $\bar{\phi}=(\phi+\phi'_1+\phi'_2)/3$.

\textbf{Diversity-Enhanced Supervised Contrastive Learning.} 
Mixed data augmentation generates diverse augmented versions of input data through random transformations and mixing strategies, enhancing the robustness and uncertainty calibration of local models. However, traditional training methods may not fully exploit this diverse information, limiting improvements in model robustness and generalization. To address this, we introduce a supervised contrastive learning strategy~\cite{nips2020supcon} that leverages augmented data to explicitly encourage the model to learn more discriminative and robust feature representations. This process enhances the generalizability of local models to unseen corruptions by exploiting diverse augmentations.

We observe that directly applying mixed data augmentation to supervised contrastive learning can be counterproductive, potentially degrading model performance, as shown in Table~\ref{tab:ablation_dcl}. This may be due to the complex nature of mixed augmentation, which can cause structural distortions in images~\cite{tpami2022clsa}, thereby undermining the effectiveness of supervised contrastive learning.
Specifically, mixed data augmentation, while designed to simulate real-world corruptions, can sometimes create overly complex transformations that distort important structural information in images. Supervised contrastive learning uses label information to learn robust and discriminative feature representations, encouraging images of the same class to cluster in feature space while separating those of different classes. However, when the augmented images contain distortions that obscure the essential features of a specific class, the model struggles to learn accurate and discriminative representations. Consequently, the complex augmented images from mixed data augmentation may undermine the advantages of supervised contrastive learning, potentially degrading performance. 

To effectively learn discriminative features from augmented images obtained by mixed data augmentation, we design a diversity-enhanced supervised contrastive learning strategy. This strategy allows local models to better utilize complex augmented data and focus on class-specific generalized features invariant to various corruptions. 
Specifically, given a batch of $B_k$ training data $D_k^b\in D_k$, where $|D_k^b|=B_k$ and $D_k^b=\{(x_i^k,y_i^k)\}_{i=1}^{B_k}$. 
For each input image $x_i\in D_k^b$, we generate a complex augmented version $x'_i$ through the randomly mixed augmentation strategy and a simple augmented version $x''_i$ through standard augmentation techniques (such as random cropping, color jittering, gaussian blur, grayscaling, and flipping, \etc). 
All version of the samples $x=\{(x_i,x'_i,x''_i)|x_i\in D_k^b\}$ are fed into the local feature extractor $f_k(u_k)$ to obtain their feature representations $f^k=f_k(x,u_k)$.

To mitigate the potential failure of supervised contrastive learning due to structural distortions in complex augmented images, we utilize simple augmented images as intermediaries. We create a multiview batch $I$ containing only original and simple augmented data, where $|I|=2B_k$. Within $I$, we form positive pairs by matching images of the same category. For each image $x_i\in I$, we define its positive sample set as $P_i={x_p|x_p\in A_i,y_p=y_i}$, where $A_i=I \backslash \{x_i\}$ denotes the set of all images in the batch $I$ except the image $x_i$. Then, we adopt the general supervised contrastive loss to bring the feature representations of positive pairs close in feature space and separate the feature representations of negative pairs, which can be expressed as:  
\begin{equation}
\begin{aligned}
\ell_c = \sum_{x_i\in I} \frac{-1}{|P_i|} \sum_{x_p \in P_i}\log \frac{\exp(f_i^k \cdot f_p^k / \tau_c)}{\sum_{x_a \in A_i} \exp(f_i^k \cdot f_a^k / \tau_c)},
\end{aligned}
\end{equation}
where $\tau_c$ is the temperature scaling parameter of supervised contrastive loss. In this way, the models effectively mine robust and discriminative feature representations.

Complex model augmentation provides diverse information with randomness and complexity, enabling the models to learn more general and robust feature representations. Therefore, fully exploring the effective information contained in the complex augmented images is crucial to improving the robustness of the models. 
However, through experiments, we found that directly forcing the models to learn from the complex augmented images using conventional supervised contrastive learning methods does not improve performance and may even cause significant degradation. The structural distortions introduced by complex augmentations can obscure essential features, making it difficult for the models to learn meaningful representations. Therefore, we adopt a softer approach that uses the similar probability distribution of simple augmented images as a supervisory signal to guide the learning of the complex augmented images. Specifically, we obtain the similarity probability distribution of the simple augmented version $x''_i$ by calculating the probability that its feature representation is similar in all other images in the batch $I$:
\begin{equation}
\begin{aligned}
p(x_j|x''_i) = \frac{\exp(f_{i''}^k \cdot f_j^k / \tau_d)}{\sum_{x_a \in A''_i} \exp(f_{i''}^k \cdot f_a^k / \tau_d)}.
\end{aligned}
\end{equation}
Here $A''_i=I\backslash\{x''_i\}$ and $x_j\in A''_i$. \blue{When $x_j\in P''_i$, where $P''_i=\{x_p|x_p\in A''_i,y_p=y_i\}$, $p(x_j|x''_i)$ represents positive likelihood; otherwise it represents negative likelihood.} $\tau_d$ denotes the temperature scaling parameter that controls the smoothness of the similarity probability distribution. Similarly, the similar probability distribution of the complex augmented version $x'$ can be expressed as:
\begin{equation}
\begin{aligned}
p(x_j|x'_i) = \frac{\exp(f_{i'}^k \cdot f_j^k / \tau_d)}{\sum_{x_a \in A''_i} \exp(f_{i'}^k \cdot f_a^k / \tau_d)}.
\end{aligned}
\end{equation}
Then, the regularization loss term $\ell_d$ is introduced to minimize the KL divergence between the similarity probability distributions of the complex augmented image and the simple augmented image, which can be represented as:
\begin{equation}
\begin{aligned}
\ell_d=\sum\nolimits_{i=1}^{N_k}p(x_j|x''_i)\log\frac{p(x_j|x''_i)}{p(x_j|x'_i)}.
\end{aligned}
\end{equation}
The regularized learning process indirectly aligns complex augmented data representations, fully exploiting their rich information and effectively improving model robustness.

In the local update phase, the clients update the local models with their private datasets, so that the local models can sufficiently learn the local information and prevent them from forgetting local knowledge after several rounds of communication. However, the client may repeatedly learn corrupted knowledge and optimize in the wrong direction, which will eventually lead to non-convergence of the local model. Thus, we utilize a novel $\ell_{local}$ as the loss function of the local update, and the local update process of client $c_k$ can be formulated as:
\begin{equation}\label{eq:localupdate}
\begin{aligned}
\theta_k^{t_l}\leftarrow&\theta_k^{t_l-1}-\lambda\nabla_\theta\ell_{local}(\phi(\tilde{x}^k;\theta_k^{t_l-1}),y^k),\\
&\ell_{local}=\ell_{ce}+\mu\ell_{jsd}+\ell_c+\gamma\ell_{d},
\end{aligned}
\end{equation}
where $\lambda$ denotes the local learning rate and $t_l\in [0,T_l]$ represents the $t_l$-th local update epoch. $\ell_{ce}$ denotes the cross-entropy loss. $\mu$ and $\gamma$ control the strength of the JS consistency constraint and the regularization constraint strength in diversity-enhanced supervised contrastive learning, respectively.

\subsection{Asymmetric Heterogeneous Federated Learning}\label{sec:colupdate}
To facilitate robust communication of heterogeneous models, we develop a general asymmetric heterogeneous FL framework, which is an improved version of the previous heterogeneous FL framework~\cite{iccv2023aughfl}. The framework focuses on optimizing communication between heterogeneous models, enabling efficient knowledge sharing despite the structural differences between client models. In addition, to promote more efficient and resilient inter-client communication and avoid the adverse effects of merging low-quality information from unreliable sources, we ensure that clients selectively learn one-way from others.

The private datasets contain proprietary information that is protected and non-exchangeable. However, unlabeled and irrelevant public data is generally readily available. The output distribution of public data on local models can reflect the discriminative capabilities of the models. Consequently, the output distribution of public datasets can be leveraged as shared information to facilitate communication between heterogeneous clients without leaking any raw private data or local model design details. \blue{Crucially, the public dataset does not need to be relevant to the tasks of clients, as its primary role is to serve as a common reference point for all participating clients. Collaborative learning between clients is promoted by capturing the relative differences in the output distributions of local models on the same public data, regardless of their local specific tasks. By leveraging irrelevant public data, which are more readily accessible and usable, we minimize potential bias and reduce the burden and privacy risk of collecting task-specific public data, making RAHFL more robust and widely applicable.}

Specifically, the client $c_k$ expresses local knowledge through the output class distribution $\phi_k(D_0,\theta_k)$ computed by the local model $\phi_k$ on public dataset $D_0$. We implement collaborative learning with KL divergence, a metric measuring the distance between two probability distributions. The discrepancy between the output class distribution from client $c_q$ to client $c_p$ is expressed as:
\begin{equation}\label{eq:hfl_kl}
\begin{aligned}
\ell_{kl}(p||q)=\sum\nolimits_{i=1}^{N_0}p(x_i^0)\log\frac{p(x_i^0)}{q(x_i^0)},
\end{aligned}
\end{equation}
where $p(x_i^0)=\phi_p(x_i^0,\theta_p)$ and $q(x_i^0)=\phi_q(x_i^0,\theta_q)$ indicate the output class distributions of public data $x_i^0$ on client $c_p$ and client $c_q$, respectively. Furthermore, the alignment of local knowledge distributions between clients can be achieved by narrowing the gap between their output class distributions. Therefore, we can achieve the process of acquiring knowledge from other clients by minimizing the KL divergence loss.

\blue{In the previous model heterogeneous collaborative learning approaches HFL, all clients participate in the federated learning process and perform simultaneous and symmetric knowledge exchange. However, this symmetric learning approach can lead to the propagation of low-quality information, especially from poorly performing clients. Incorporating knowledge from poorly performing clients degrades the overall learning efficiency, slows convergence, and decreases the final model accuracy. In addition, the disparity in model performance and learning efficiency among heterogeneous models across clients can cause some clients to achieve optimal states earlier than others. These clients will exit the process of learning from others to avoid unnecessary updates. Furthermore, the presence of data corruption further exacerbates these problems.}

Therefore, AsymHFL allows clients to selectively and asymmetrically learn from other clients in each round of collaborative learning. In this approach, clients only learn from clients that outperform their own. This asymmetric learning strategy ensures that the transmitted information is high-quality and beneficial, thereby preventing the propagation of low-quality knowledge and promoting more reliable and effective knowledge transfer. Specifically, we define $M\in \mathbb{R}^{(K\times K)}$ as a knowledge transfer matrix to control the asymmetric learning relationship between clients, which can be expressed as:
\begin{equation}
\begin{aligned}
M^{t_c}=
\begin{Bmatrix}
 0 &  M_{12}^{t_c}& \cdots & M_{1K}^{t_c}\\
 M_{21}^{t_c} &  0& \cdots & M_{2K}^{t_c}\\
 \vdots &  \vdots& \ddots & \vdots\\
 M_{K1}^{t_c} &  M_{K2}^{t_c}& \cdots & 0
\end{Bmatrix},
\end{aligned}
\end{equation}
with
\begin{equation}
\begin{aligned}
M_{pq}^{t_c}=
\begin{cases}
 0,&\text{ if } ACC(\phi_p) > ACC(\phi_q)\\
 1,&\text{ if } ACC(\phi_p)\le ACC(\phi_q)
\end{cases},
\end{aligned}
\end{equation}
where $t_c\in [0,t_c]$ represents the $t_c$-th collaborative update round, and $M_{pq}^{t_c}\in\{0,1\}$ represents whether client $c_p$ learns from client $c_q$ in the $t_c$-th collaborative update round. If client $c_p$ learns from client $c_q$ in $t_c$-th collaborative learning round, then $M_{pq}=1$, otherwise $M_{pq}=0$. The decision for a client to learn from another is dynamically determined based on the comparative performance of their local models, which is evaluated by the classification accuracy $ACC(x_t,\phi)$ of its local model on a part of the generic test dataset $x_t$. The knowledge transfer matrix $M$ can be dynamically updated in subsequent rounds based on the evolving performance of the client models, continuously optimizing the knowledge transfer process and ensuring that the clients always learn from the effective source. Such a dynamic update mode promotes continuous improvement of all clients rather than static convergence to a single optimal model. \blue{In symmetric knowledge transfer, all client pairs need to calculate knowledge distribution differences. In contrast, our asymmetric approach only requires calculations for selected clients, potentially reducing the computational load of collaborative learning by up to 50\%.}

\definecolor{mygray}{gray}{.9}
\begin{table*}[!h]\small
\setlength{\abovecaptionskip}{0cm} 
\centering
\caption{Ablation experiment with corruption rate $\xi=0$ on private dataset, $\theta_k$ represents the local model of client $c_k$.}
\label{tab:ablationclean} 
\begin{tabular}{cccc||P{0.8cm}P{0.8cm}P{0.8cm}P{0.8cm}|P{0.8cm}||P{0.8cm}P{0.8cm}P{0.8cm}P{0.8cm}|P{0.8cm}}\hline
 \specialrule{.1em}{0em}{0em}
\rowcolor{mygray}   
\multicolumn{4}{c||}{Components}& \multicolumn{5}{c||}{Test on clean dataset}& \multicolumn{5}{c}{Test on random corrupted dataset} \\
\hhline{----||-----||-----}
\rowcolor{mygray}
HFL & AsymHFL  & Aug & DCL & $\theta_0$ & $\theta_1$ & $\theta_2$ & $\theta_3$& {Avg}& $\theta_0$ & $\theta_1$ & $\theta_2$ & $\theta_3$& {Avg}\\
\hhline{----||-----||-----}
 & & & & 82.63 & 82.57 & 68.58 & 78.14 & 77.98 & 68.03 & 66.21 & 56.04 & 63.23 & 63.38\\
 $\checkmark$ & & & & 81.18 & 81.29 & 71.46 & 80.50 & 78.61 & 65.75 & 65.78 & 58.08 & 64.20 & 63.45\\
 & $\checkmark$ & & & 83.89& 83.21& 75.89& 81.91& 81.23& 68.09& 66.13& 62.57& 66.94& 65.93\\
 & $\checkmark$ & $\checkmark$ & & 85.37& 84.80& 77.71& 83.79& 82.92& 77.31& 77.87& 72.77& 75.41& 75.84\\
 & $\checkmark$ & $\checkmark$ & $\checkmark$ & \textbf{87.00}& \textbf{86.75}& \textbf{78.70}& \textbf{85.16}& \textbf{84.40}& \textbf{79.58}& \textbf{78.25}& \textbf{73.01}& \textbf{78.89}& \textbf{77.43}\\
 \specialrule{.1em}{0em}{0em}
 \end{tabular}
  \vspace{-2mm}
\end{table*}

\definecolor{mygray}{gray}{.9}
\begin{table*}[!h]\small
\setlength{\abovecaptionskip}{0cm} 
\centering
\caption{Ablation experiment with corruption rate $\xi=0.5$ on private dataset, $\theta_k$ represents the local model of client $c_k$.}
\label{tab:ablationhalfrandom} 
\begin{tabular}{cccc||P{0.8cm}P{0.8cm}P{0.8cm}P{0.8cm}|P{0.8cm}||P{0.8cm}P{0.8cm}P{0.8cm}P{0.8cm}|P{0.8cm}}\hline
 \specialrule{.1em}{0em}{0em}
\rowcolor{mygray}   
\multicolumn{4}{c||}{Components}& \multicolumn{5}{c||}{Test on clean dataset}& \multicolumn{5}{c}{Test on random corrupted dataset} \\
\hhline{----||-----||-----}
\rowcolor{mygray}
HFL & AsymHFL  & Aug & DCL & $\theta_0$ & $\theta_1$ & $\theta_2$ & $\theta_3$& {Avg}& $\theta_0$ & $\theta_1$ & $\theta_2$ & $\theta_3$& {Avg}\\
\hhline{----||-----||-----}
 & & & & 68.75 & 68.46 & 57.14 & 57.41 & 62.94 & 64.03 & 64.75 & 50.86 & 52.40 & 58.01\\
 $\checkmark$ & & & & 61.37 & 63.71 & 56.08 & 59.34 & 60.13 & 57.19 & 60.07 & 50.79 & 56.25 & 56.08 \\
 & $\checkmark$ & & & 80.19& 82.58& 74.09& 79.67& 79.13& 76.68& 75.69& 70.58& 74.81& 74.44\\
 & $\checkmark$ & $\checkmark$ & & 81.67& 81.67& 76.56& 80.66& 80.14& 78.94& 76.78& \textbf{74.25}& 76.32& 76.57\\
 & $\checkmark$ & $\checkmark$ & $\checkmark$ & \textbf{82.46}& \textbf{83.58}& \textbf{76.57}& \textbf{81.70}& \textbf{81.08}& \textbf{79.29}& \textbf{80.62}& 73.85& \textbf{77.6}1& \textbf{77.84}\\
 \specialrule{.1em}{0em}{0em}
 \end{tabular}
  \vspace{-2mm}
\end{table*}

\definecolor{mygray}{gray}{.9}
\begin{table*}[!h]\small
\setlength{\abovecaptionskip}{0cm} 
\centering
\caption{Ablation experiment with corruption rate $\xi=1$ on private dataset, $\theta_k$ represents the local model of client $c_k$.}
\label{tab:ablationrandom} 
\begin{tabular}{cccc||P{0.8cm}P{0.8cm}P{0.8cm}P{0.8cm}|P{0.8cm}||P{0.8cm}P{0.8cm}P{0.8cm}P{0.8cm}|P{0.8cm}}\hline
 \specialrule{.1em}{0em}{0em}
\rowcolor{mygray}   
\multicolumn{4}{c||}{Components}& \multicolumn{5}{c||}{Test on clean dataset}& \multicolumn{5}{c}{Test on random corrupted dataset} \\
\hhline{----||-----||-----}
\rowcolor{mygray}
HFL & AsymHFL  & Aug & DCL & $\theta_0$ & $\theta_1$ & $\theta_2$ & $\theta_3$& {Avg}& $\theta_0$ & $\theta_1$ & $\theta_2$ & $\theta_3$& {Avg}\\
\hhline{----||-----||-----}
 & & & & 65.58 & 67.01 & 56.00 & 57.11 & 61.43 & 62.47 & 63.88 & 52.57 & 53.49 & 58.10 \\
 $\checkmark$ & & & & 62.02 & 62.24 & 58.51 & 59.84 & 60.65 & 59.22 & 59.88 & 55.00 & 56.74 & 57.71 \\
 & $\checkmark$ & & & 79.07& 80.37& 71.69& 79.47& 77.65& 76.27& 74.50& 69.42& 75.67& 73.96\\
 & $\checkmark$ & $\checkmark$ & & \textbf{81.76}& 82.04& \textbf{76.22}& 79.59& 79.91& 77.80& 78.13& 71.61& 76.78& 76.08\\
 & $\checkmark$ & $\checkmark$ & $\checkmark$ & 81.15& \textbf{82.58}& 76.14& \textbf{80.67}& \textbf{80.13}& \textbf{78.42}& \textbf{80.09}& \textbf{73.60}& \textbf{78.24}& \textbf{77.59}\\
 \specialrule{.1em}{0em}{0em}
 \end{tabular}
  \vspace{-2mm}
\end{table*}

\definecolor{mygray}{gray}{.9}
\begin{table*}[t]\small
\centering
 \caption{Ablation study of DCL. The average test accuracy of local models is demonstrated. The term 'w/o SCL' refers to the result of RAHFL without any supervised contrastive learning strategies, i.e., AsymHFL+Aug. The variable $\xi$ denotes the corruption rate. 'Clean' and 'Corrupted' represent testing on clean dataset and testing on corrupted dataset, respectively.
 }
   \vspace{-2mm}
\begin{tabular}{c||P{1.5cm}P{1.5cm}||P{1.5cm}P{1.5cm}||P{1.5cm}P{1.5cm}}\hline
  \specialrule{.1em}{0em}{0em}
\rowcolor{mygray}
   & \multicolumn{2}{c||}{$\xi=0$}  & \multicolumn{2}{c||}{$\xi=0.5$} & \multicolumn{2}{c}{$\xi=1$}\\\hhline{~||--||--||--}
\rowcolor{mygray}    \multirow{-2}{*}{Method} &{Clean} &{Corrupted} &{Clean} &{Corrupted}  &{Clean} &{Corrupted}  \\\hhline{-||--||--||--}
 w/o SCL&  82.92& 75.84& 80.14& 76.57& 79.91& 76.08 \\
 +SupCon&  83.31& 74.37& 79.88& 76.35& 79.44& 76.18 \\
 +DCL&  \textbf{84.40}& \textbf{77.43}& \textbf{81.08}& \textbf{77.84}& \textbf{80.13}& \textbf{77.59} \\\hline
  \specialrule{.1em}{0em}{0em}
 \end{tabular}
  \vspace{-3mm}
\label{tab:ablation_dcl}
\end{table*}

In the collaborative update phase, all clients calculate their local knowledge distribution using the local model and the public dataset. Then, the knowledge transfer matrix $M$ is constructed based on the client model performance. Finally, the knowledge transfer matrix $M$ enables client $c_k$ to selectively learn from the knowledge distribution of clients with superior performance. Therefore, the loss of the client $c_k$ in the $t_c$-th collaborative round can be formulated as:
\begin{equation}\label{eq:hfl_kl2}
\begin{aligned}
\ell_{col}^{k,t_c}=\sum\nolimits_{i=1}^{K}M_{ki}^{t_c}\cdot \ell_{kl}(\phi_0^{i,t_c}||\phi_0^{k,t_c}),
\end{aligned}
\end{equation}
where $\phi_0^{k,t_c}=\phi_k(D_0,\theta_k^{t_c})$ is the prediction distribution of the local model $\phi_k$ on public dataset $D_0$, and $\phi_0^{i,t_c}$ is similar. In this way, each client $c_k$ selectively learns from others by fitting the output class distributions of other clients. 

Compared with symmetric learning methods, in which all clients exchange information indiscriminately, our asymmetric FL framework significantly reduces the risk of integrating low-quality information or corrupted feedback. This ensures more reliable and effective knowledge transfer between heterogeneous models and improves the stability of model updates. Additionally, it accelerates the convergence of FL systems by focusing on high performance models.

\section{Experimental}
\subsection{Experimental Setup}
\noindent\textbf{Datasets and Models.} Following recent works~\cite{cvpr2022rhfl,iclr2020augmix}, our experiments utilize the Cifar-10-C~\cite{iclr2019corruption} and Cifar-100~\cite{2009cifar} datasets, due to their importance in the research on data corruption learning and image classification tasks. Both Cifar-10~\cite{2009cifar} and Cifar-100 contain 60,000 color images of size $32\times32$, with 50,000 images designated for training and 10,000 for testing. Cifar-10-C is derived from Cifar-10 by applying common visual corruptions. In our setup, Cifar-10-C is randomly partitioned among clients as their private datasets, while a subset of Cifar-100 is allocated as the public dataset on the server. To accommodate the heterogeneous model scenario, we assign four
different local models to four clients: ResNet10~\cite{cvpr2016resnet}, ResNet12~\cite{cvpr2016resnet}, ShuffleNet~\cite{cvpr2018shufflenet} and MobileNetv2~\cite{cvpr2018mobilenetv2}. 

\noindent\textbf{Corruption Patterns.}
To closely mimic real-world data corruption, we construct corruption patterns for FL following Hendrycks \etal~\cite{iclr2019corruption}. Cifar-10-C consists of 15 types of corruption, categorized into Noise, Blur, Weather, and Digital, each with five severity levels. Corrupted images are generated by randomly sampling the corruption types and the severity levels from a uniform distribution. The private dataset of each client is subjected to varying corruption rates, corruption types, and severity levels, and different samples within a private dataset can exhibit distinct corruption patterns.

\noindent\textbf{Baselines.} We compare RAHFL with several SOTA methods to demonstrate its effectiveness in heterogeneous model scenarios. Specifically, RAHFL is compared with FedMD~\cite{nips2019fedmd}, FedDF~\cite{nips2020feddf}, KT-pFL~\cite{nips2022ktpfl}, RHFL~\cite{cvpr2022rhfl}, FCCL~\cite{cvpr2022fccl}, FedProto~\cite{aaai2022fedproto} and AugHFL~\cite{iccv2023aughfl} under the consistent setting. FedMD relies on average class scores from client models on a public dataset for communication. FedDF is a distillation-based framework for robust federated model fusion using unlabeled or synthetic data. KT-pFL uses a knowledge coefficient matrix for personalized knowledge transfer. FedProto employs a prototype learning strategy to address data and model heterogeneity in FL. RHFL addresses label noise and model heterogeneity within a unified framework. FCCL uses a cross-correlation matrix for collaborative learning. AugHFL designs a robust adaptive re-weighted communication strategy based on enhanced consistency constraints.
The original experimental settings of these algorithms are different, so we implement the main ideas of these algorithms in our framework for a fair comparison.

\definecolor{mygray}{gray}{.9}
\begin{table*}[!h]\small
\setlength{\abovecaptionskip}{0cm} 
\centering
\caption{Compare with the SOTA methods with corruption rate $\xi=0$ on private dataset, $\theta_k$ represents the local model of the client $c_k$.}
\label{tab:sotaclean} 
\begin{tabular}{P{2.3cm}||P{1.1cm}P{1.1cm}P{1.1cm}P{1.1cm}|P{1.1cm}||P{1.1cm}P{1.1cm}P{1.1cm}P{1.1cm}|P{1.1cm}}
\hline
\specialrule{.1em}{0em}{0em}
\rowcolor{mygray}
    & \multicolumn{5}{c||}{Test on clean dataset} & \multicolumn{5}{c}{Test on random corrupted dataset}\\
\hhline{*1>{\arrayrulecolor[gray]{.9}}->{\arrayrulecolor{black}}||----|-||----|-}
\rowcolor{mygray}
    \multirow{-2}{*}{Model} & $\theta_0$ & $\theta_1$ & $\theta_2$ & $\theta_3$ & Avg & $\theta_0$ & $\theta_1$ & $\theta_2$ & $\theta_3$ & Avg\\
\hhline{-||----|-||----|-}
    Baseline & 82.63 & 82.57 & 68.58 & 78.14 & 77.98 & 68.03 & 66.21 & 56.04 & 63.23 & 63.38\\
    FedMD\cite{nips2019fedmd} & 82.86 & 83.06 & 73.02 & 80.74 & 79.92 & 68.70 & 68.28 & 61.01 & 67.53 & 66.38\\
    FedDF\cite{nips2020feddf} & 82.40 & 82.40 & 73.55 & 78.48 & 79.21 & 65.54 & 67.00 & 59.39 & 63.36 & 63.82\\
    KT-pFL\cite{nips2022ktpfl} & 84.73 & 85.07 & 72.74 & 81.06 & 80.90 & 68.76 & 68.21 & 60.58& 66.48& 66.01\\
    RHFL\cite{cvpr2022rhfl} & 82.30 & 82.83 & 71.72 & 77.55 & 78.60 & 66.32 & 62.28 & 57.65 & 62.94 & 62.30\\
    FCCL\cite{cvpr2022fccl} & 83.26 & 83.07 & 73.69 & 81.53 & 80.39 & 69.55 & 67.78 & 60.57 & 68.25 & 66.54\\
    FedProto\cite{aaai2022fedproto} & 84.84& 85.52& 72.35& 81.57& 81.07& 68.62& 67.69& 60.20& 66.61& 65.78\\
\hhline{-||----|-||----|-}
    AugHFL~\cite{iccv2023aughfl} & 79.86 & 81.45 & 70.67 & 79.47 & 77.86 & 73.78 & 74.46 & 64.03 & 72.55 & 71.21\\
    RAHFL & \textbf{87.00}& \textbf{86.75}& \textbf{78.70}& \textbf{85.16}& \textbf{84.40}& \textbf{79.58}& \textbf{78.25}& \textbf{73.01}& \textbf{78.89}& \textbf{77.43}\\
\hline
\specialrule{.1em}{0em}{0em}
\end{tabular}
  \vspace{-2mm}
\end{table*}

\begin{table*}[!h]\small
\setlength{\abovecaptionskip}{0cm}
\centering
\caption{Compare with the SOTA methods with corruption rate $\xi=0.5$ on private dataset, $\theta_k$ represents the local model of the client $c_k$.}
\label{tab:sotahalfrandom}
\begin{tabular}{P{2.3cm}||P{1.1cm}P{1.1cm}P{1.1cm}P{1.1cm}|P{1.1cm}||P{1.1cm}P{1.1cm}P{1.1cm}P{1.1cm}|P{1.1cm}}
\hline
\specialrule{.1em}{0em}{0em}
\rowcolor{mygray}
    & \multicolumn{5}{c||}{Test on clean dataset} & \multicolumn{5}{c}{Test on random corrupted dataset}\\
\hhline{*1>{\arrayrulecolor[gray]{.9}}->{\arrayrulecolor{black}}||----|-||----|-}
\rowcolor{mygray}
    \multirow{-2}{*}{Model} & $\theta_0$ & $\theta_1$ & $\theta_2$ & $\theta_3$ & Avg & $\theta_0$ & $\theta_1$ & $\theta_2$ & $\theta_3$ & Avg\\
\hhline{-||----|-||----|-}
    Baseline & 68.75 & 68.46 & 57.14 & 57.41 & 62.94 & 64.03 & 64.75 & 50.86 & 52.40 & 58.01\\
    FedMD\cite{nips2019fedmd} & 64.25 & 65.30 & 55.97 & 58.58 & 61.03 & 59.66 & 61.28 & 51.66 & 54.56 & 56.79\\
    FedDF\cite{nips2020feddf} & 61.72 & 63.51 & 57.29 & 57.89 & 60.10 & 58.09 & 59.57 & 53.15 & 54.35 & 56.29\\
    KT-pFL\cite{nips2022ktpfl} & 74.68& 75.85& 62.26& 69.16& 70.49& 71.68& 71.63& 58.07& 65.54& 66.73\\
    RHFL\cite{cvpr2022rhfl} & 58.05 & 59.47 & 51.13 & 55.07 & 55.93 & 62.42 & 63.46 & 56.78 & 59.81 & 60.62\\
    FCCL\cite{cvpr2022fccl} & 63.75 & 62.62 & 58.43 & 59.78 & 61.15 & 59.91 & 59.03 & 53.71 & 56.24 & 57.22\\
    FedProto\cite{aaai2022fedproto} & 74.18& 76.13& 64.49& 69.92& 71.18& 71.52& 71.79& 59.94& 65.57& 67.21\\
\hhline{-||----|-||----|-}
    AugHFL~\cite{iccv2023aughfl} & 76.22 & 76.50 & 66.66 & 73.31 & 73.17 & 71.96 & 71.26 & 61.28 & 69.58 & 68.52\\
    RAHFL & \textbf{82.46}& \textbf{83.58}& \textbf{76.57}& \textbf{81.70}& \textbf{81.08}& \textbf{79.29}& \textbf{80.62}& \textbf{73.85}& \textbf{77.61}& \textbf{77.84} \\
\hline
\specialrule{.1em}{0em}{0em}
\end{tabular}
  \vspace{-2mm}
\end{table*}

\begin{table*}[!h]\small
\setlength{\abovecaptionskip}{0cm}
\centering
\caption{Compare with the SOTA methods with corruption rate $\xi=1$ on private dataset, $\theta_k$ represents the local model of the client $c_k$.}
\label{tab:sotarandom}
\begin{tabular}{P{2.3cm}||P{1.1cm}P{1.1cm}P{1.1cm}P{1.1cm}|P{1.1cm}||P{1.1cm}P{1.1cm}P{1.1cm}P{1.1cm}|P{1.1cm}}
\hline
\specialrule{.1em}{0em}{0em}
\rowcolor{mygray}
    & \multicolumn{5}{c||}{Test on clean dataset} & \multicolumn{5}{c}{Test on random corrupted dataset}\\
\hhline{*1>{\arrayrulecolor[gray]{.9}}->{\arrayrulecolor{black}}||----|-||----|-}
\rowcolor{mygray}
    \multirow{-2}{*}{Model} & $\theta_0$ & $\theta_1$ & $\theta_2$ & $\theta_3$ & Avg & $\theta_0$ & $\theta_1$ & $\theta_2$ & $\theta_3$ & Avg\\
\hhline{-||----|-||----|-}
    Baseline & 65.58 & 67.01 & 56.00 & 57.11 & 61.43 & 62.47 & 63.88 & 52.57 & 53.49 & 58.10 \\
    FedMD\cite{nips2019fedmd} & 62.11 & 63.74 & 56.57 & 58.12 & 60.14 & 59.11 & 60.58 & 52.51 & 55.39 & 56.90 \\
    FedDF\cite{nips2020feddf} & 61.66 & 63.11 & 58.39 & 58.06 & 60.31 & 58.87 & 59.45 & 55.44 & 55.31 & 57.27 \\
    KT-pFL\cite{nips2022ktpfl} & 73.39& 73.66& 60.68& 67.13& 68.71& 70.88& 71.26& 57.07& 64.26& 65.87\\
    RHFL\cite{cvpr2022rhfl} & 62.55 & 64.21 & 57.57 & 58.14 & 60.62 & 58.14 & 59.79 & 54.75 & 55.77 & 57.11 \\
    FCCL\cite{cvpr2022fccl} & 62.60 & 63.64 & 58.60 & 59.71 & 61.14 & 59.36 & 59.55 & 55.82 & 56.92 & 57.91 \\
    FedProto\cite{aaai2022fedproto} & 73.73& 74.75& 62.70& 67.47& 69.66& 70.84& 71.86& 59.76& 64.49& 66.74\\
\hhline{-||----|-||----|-}
    AugHFL~\cite{iccv2023aughfl} & 76.98 & 77.82 & 65.79 & 74.03 & 73.66 & 73.65 & 74.04 & 61.50 & 70.92 & 70.03 \\
    RAHFL & \textbf{81.15}& \textbf{82.58}& \textbf{76.14}& \textbf{80.67}& \textbf{80.13}& \textbf{78.42}& \textbf{80.09}& \textbf{73.60}& \textbf{78.24}& \textbf{77.59} \\
\hline
\specialrule{.1em}{0em}{0em}
\end{tabular}
  \vspace{-2mm}
\end{table*}

\noindent\textbf{Implementation Details.} Initially, the local model of each client is pre-trained for $40$ epochs with its private dataset. The size of private datasets and the public dataset is specified as $N_k=10,000$ and $N_0=5,000$ respectively. For effective FL, all clients perform $T_c=40$ rounds of collaborative updates. The number of local learning epochs is adaptively set to $T_l=\max(\lfloor\frac{N_0}{N_K}\rfloor,1)$ to balance local knowledge and information from others. Furthermore, we employ the Adam~\cite{iclr2014adam} optimizer with an initial learning rate of $\lambda=0.001$ and a batch size of 256. The number of augmentation operation sequences $\mathcal{S}$ is set to $3$. The hyperparameters $\mu$ and $\gamma$ are $12$ and $1$ respectively. The temperature parameters $\tau_c$ and $\tau_d$ both set to $0.2$. \blue{Besides, we follow existing contrastive learning methods~\cite{cvpr2020moco,nips2020supcon,tpami2022clsa}, which sequentially apply RandomResizedCrop, ColorJitter, RandomGrayscale, GaussianBlur, and RandomHorizontalFlip strategies to generate simple augmented data in DCL. RandomResizedCrop and RandomHorizontalFlip change the spatial features of the image, while ColorJitter and RandomGrayscale modify its color properties. GaussianBlur introduces subtle changes to the sharpness of images. This combination ensures that augmented images retain enough of the original structure to be recognizable, while still providing diverse variations for robust feature learning. These augmentations are applied probabilistically, meaning not all of them are applied to every image.}
Since this paper focuses on the data corruption problem in FL, we consider three cases with corruption rates $\xi$ of 0, 0.5, and 1, representing clean, half corrupted, and fully corrupted private datasets, respectively. Each client $c_k$ randomly selects $N_K$ samples from Cifar-10-C with random corruption. The effectiveness of our method is tested on clean and randomly corrupted datasets.

\subsection{Ablation Study}
We train on private datasets with three corruption rates $\xi=0,0.5,1$, and evaluate the components on both completely clean and completely random corrupted datasets. This dual evaluation approach allows us to comprehensively assess the robustness and effectiveness of our method under various complex scenarios.

\noindent\textbf{Effectiveness of AsymHFL.} 
In Table~\ref{tab:ablationclean}\&\ref{tab:ablationhalfrandom}\&\ref{tab:ablationrandom}, the first row indicates that no components are added, \ie independently trained local models without any collaborative learning. The HFL component involves standard symmetric knowledge sharing between clients, while AsymHFL refers to our asymmetric collaborative learning strategy. The results in Table~\ref{tab:ablationclean}\&\ref{tab:ablationhalfrandom}\&\ref{tab:ablationrandom} demonstrate that the model performance significantly improves with the addition of AsymHFL in all scenarios. Notably, when the training data is half corrupted, the addition of AsymHFL leads to a remarkable improvement, \ie results tested on the completely clean dataset increase by 16.19\%, and results tested on the completely corrupted dataset increase by 16.43\%. \blue{In the presence of data corruption, adding HFL results in lower test accuracy. This indicates that in scenarios with data corruption, general symmetric FL leads to the repeated exchange of erroneous knowledge, thereby degrading model performance.} In contrast, AsymHFL effectively prevents the transfer of low-quality information from underperforming and unreliable clients, resulting in stable model improvements. 

\noindent\textbf{Effectiveness of Aug.} 
Table~\ref{tab:ablationclean}\&\ref{tab:ablationhalfrandom}\&\ref{tab:ablationrandom} verify the effectiveness of the Aug module, which incorporates random mixed augmentation with JSD consistency loss. The experimental results indicate that Aug can effectively handle data corruption. In particular, when the models are trained with clean datasets and tested on corrupted datasets, the inclusion of Aug brings an average model accuracy improvement of 9.91\%. In the local update phase, the Aug module enhances model performance, robustness to data corruption, and generalization to unseen test data. The incorporation of JSD loss enforces a consistency constraint on the output distribution of mixed augmented images, thereby enhancing model stability.

\begin{figure*}[t]
\centering{
    \subcaptionbox{Test on clean dataset}{\includegraphics[width=0.49\linewidth]{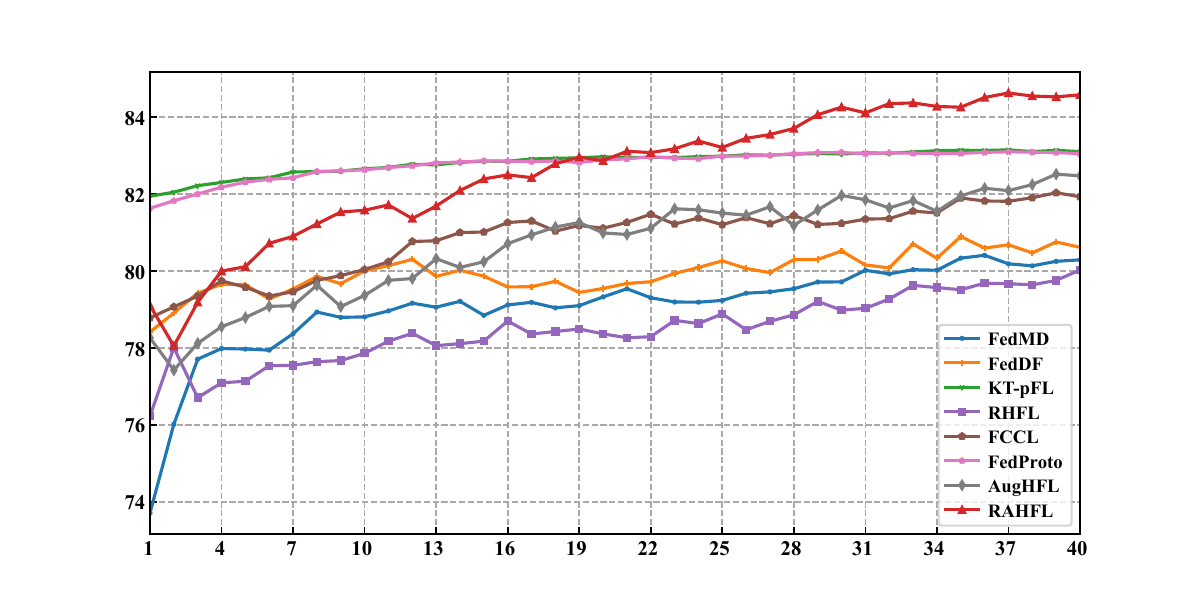}}
    \subcaptionbox{Test on corrupted dataset}
    {\includegraphics[width=0.49\linewidth]{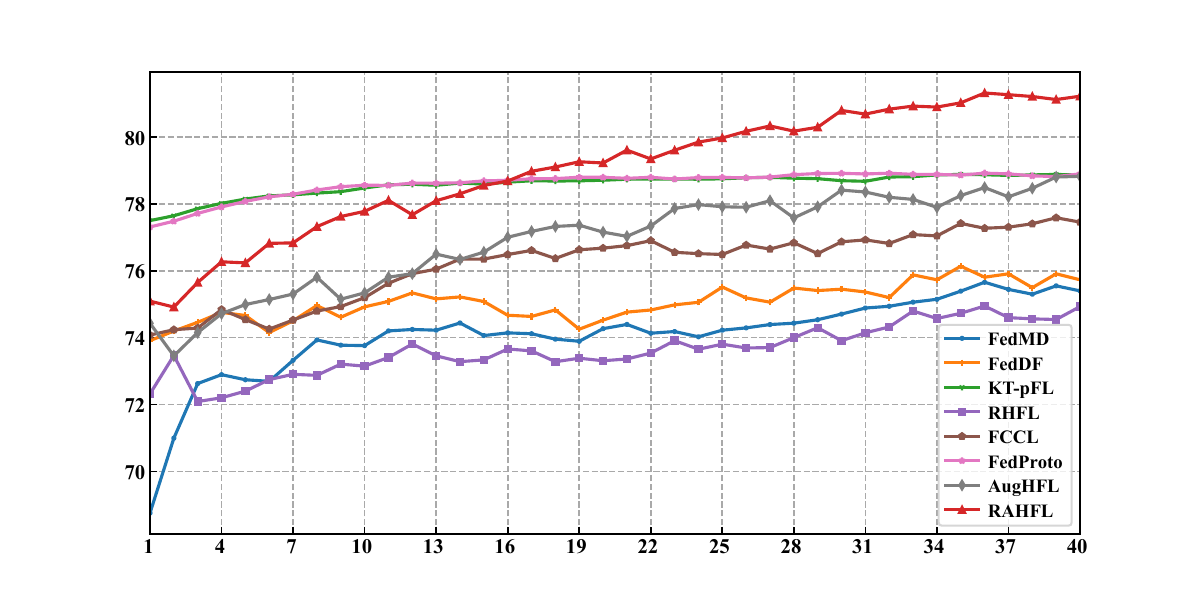}}
    \vspace{-1mm}
    \caption{Comparison of test accuracy during the FL process across homogeneous models, when the corruption rate $\xi=0.5$. The x-axis represents communication epochs, and the y-axis denotes test accuracy.}
    \label{fig:homoclient}}
  \vspace{-4mm}
\end{figure*}

\noindent\textbf{Effectiveness of DCL.} 
To demonstrate the effectiveness of the DCL module, we conducted ablation experiments to compare the model test accuracy under various settings, as shown in Table~\ref{tab:ablationclean}\&\ref{tab:ablationhalfrandom}\&\ref{tab:ablationrandom}. In Table~\ref{tab:ablation_dcl}, we compare the effectiveness of different strategies, including RAHFL without any supervised contrastive learning strategy (w/o SCL), with standard supervised contrastive learning (+SupCon), and with our proposed diversity-enhanced supervised contrastive learning (+DCL). Specifically, SupCon directly uses complex augmented data for supervised contrastive learning, and the models do not show effective improvement. DCL uses simple enhanced data as an intermediary for supervised contrastive learning, and designs a regularization mechanism to mine novel patterns in complex enhanced data. The experimental results in Table~\ref{tab:ablation_dcl} show the average test accuracy of the local models on clean and corrupted datasets under different corruption rates $\xi$. When $\xi=0$ (clean), the addition of DCL improves the test accuracy by 1.48\% on the clean dataset and 1.59\% on the corrupted dataset. As the corruption rate increases to $\xi=0.5$ and $\xi=1$, the DCL module consistently leads to better performance, with improved accuracy on both clean and corrupted datasets. These results demonstrate that the DCL module significantly enhances the robustness of the model against data corruption and effectively leverages complex augmented data to improve model performance under different levels of data corruption.

\subsection{Comparison with SOTA Methods}\label{sec:experiment_sota}
\noindent\textbf{Heterogeneous Federated Learning Settings.} 
We provide a comprehensive comparison of RAHFL with SOTA heterogeneous FL methods under various scenarios, where the models are trained on datasets with different corruption rates, and tested on both clean and randomly corrupted datasets, as shown in Table~\ref{tab:sotaclean}\&\ref{tab:sotahalfrandom}\&\ref{tab:sotarandom}. The baseline refers to the method where the clients only train locally without the FL process. The experiments demonstrate that RAHFL significantly outperforms other methods in all settings, sometimes even exceeding the current best method AugHFL by more than $9\%$ in average test accuracy.
Notably, in the ideal scenario of completely clean training and testing data, AugHFL struggles to compete with other SOTA methods, but the improved RAHFL can perform much better than the SOTA method, although this scenario is unlikely to occur in real applications. As the corruption rate in the training dataset increases from completely clean to fully corrupted, the performance of most SOTA methods degrades significantly. For instance, the average accuracy of FCCL on the clean dataset drops by 19.25\%, and the average accuracy of FedMD on the corrupted dataset drops by 9.48\%. In contrast, as the corruption rate increases from 0 to 1, the average accuracy of RAHFL on the clean dataset drops by only 4.37\%, and the average accuracy on the corrupted dataset remains largely unaffected, maintaining the highest performance among all methods. Overall, RAHFL significantly outperforms existing methods in terms of accuracy and robustness across various degrees of data corruption, demonstrating its strong resilience to data corruption in the FL environment and the effectiveness of the method in real-world applications.

\noindent\textbf{Homogeneous Federated Learning Settings.} 
As shown in Figure~\ref{fig:homoclient}, in addition to comparing RAHFL with SOTA methods in heterogeneous FL settings, we also expand the evaluation scope to homogeneous FL settings to fully demonstrate the superiority of our method in various FL scenarios. We standardized the local models of all clients to ResNet12 architecture for consistency. Figure~\ref{fig:homoclient} shows the comparison results of RAHFL with SOTA FL methods on clean and corrupted datasets at a corruption rate $\xi$ of 0.5.
The results clearly demonstrate that RAHFL maintains its advantage in homogeneous FL scenarios. When $\xi = 0.5$, RAHFL achieves an average test accuracy of 84.66\% on clean datasets and 81.36\% on corrupted datasets, outperforming other methods. Moreover, as the number of collaborative training rounds increases, The superiority of RAHFL becomes more pronounced, consistently outperforming other methods. Existing methods struggle to achieve significant performance improvements in the presence of data corruption, highlighting the robustness and effectiveness of RAHFL in maintaining stable performance in challenging environments.

\section{Conclusion}
In this paper, we introduce a novel framework RAHFL to solve the problem of robust heterogeneous FL. RAHFL effectively and efficiently mitigates the negative impact of local data corruption and the propagation of corrupted knowledge from others, significantly improving the robustness of the FL system. We proposed a diversity-enhanced supervised contrastive learning strategy, which effectively mines the rich information of complex augmented samples for supervised contrastive learning, thereby preventing performance degradation due to local data corruption. Besides, we design a general asymmetric heterogeneous FL method, which enables heterogeneous model communication while selectively transferring knowledge asymmetrically to prevent the integration of corrupt feedback. Through extensive experiments, we find that RAHFL outperforms SOTA methods in various complex FL scenarios, demonstrating its effectiveness and robustness.

\blue{\textbf{Discussion.} While RAHFL demonstrates excellent robustness in various noisy FL scenarios, it still has some limitations. RAHFL introduces additional computational processes in both the local and collaborative learning stages, which may increase the overall complexity and resource requirements of the FL system. While effective, diversity-augmented supervised contrastive learning in the local stage and knowledge transfer in the collaborative stage may increase computational overhead. In addition, while relying on an additional public dataset for collaboration between heterogeneous models is beneficial for knowledge transfer, it introduces additional requirements that may not always be easily met in all real-world scenarios. These factors may limit the applicability of our approach in resource-constrained settings or when no suitable public datasets are available. In future work, we plan to explore ways to optimize these processes and investigate alternatives that reduce the reliance on additional data, aiming to make our approach more broadly applicable and efficient.}

\bibliographystyle{IEEEtran}
\bibliography{rahfl}

\begin{IEEEbiography}[{\includegraphics[width = 1in, height = 1.25in, clip, keepaspectratio]{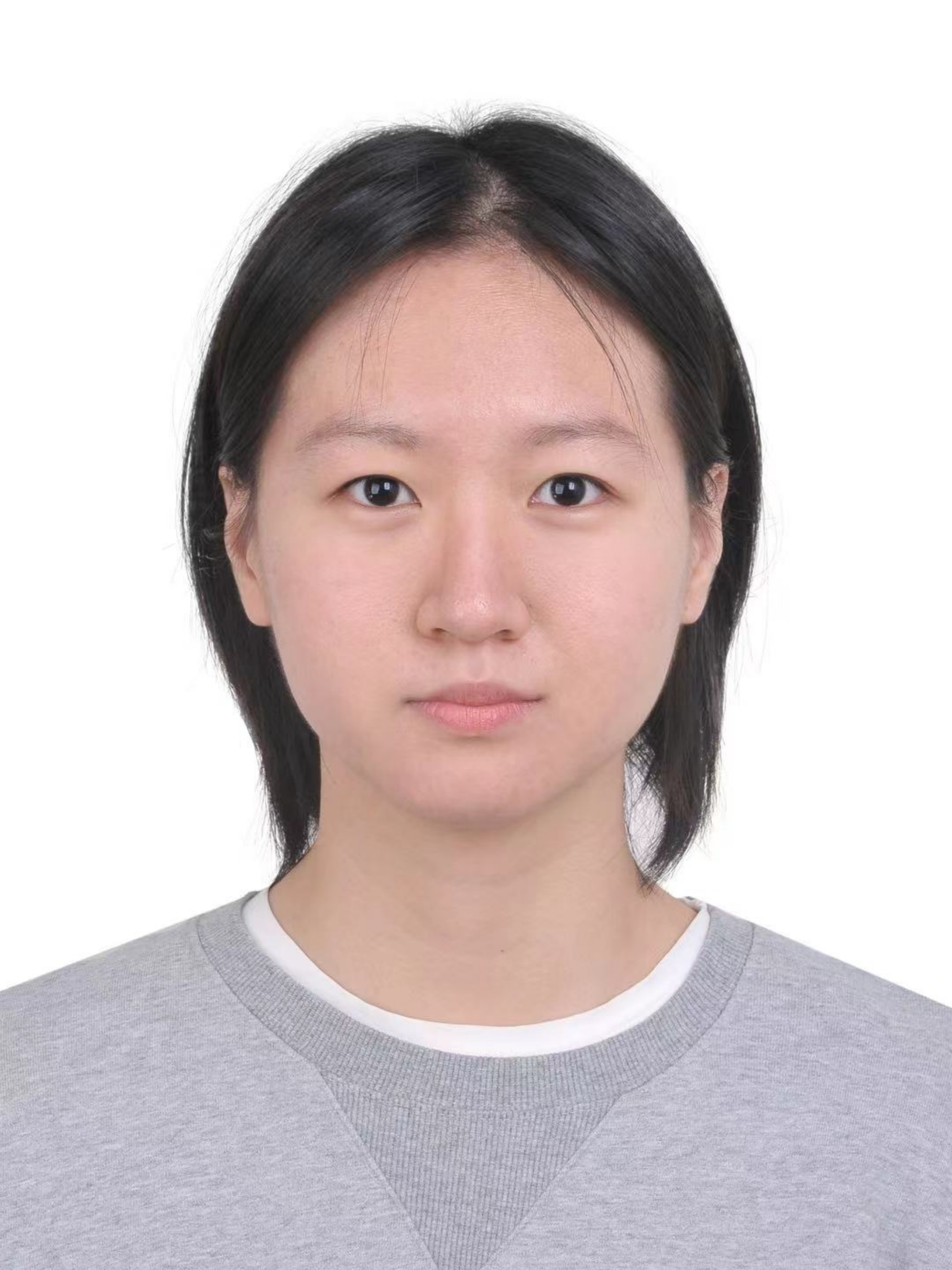}}]{Xiuwen Fang }
received the Master degree in Computer Science from Wuhan University, Wuhan, China, in 2023. She is currently a Ph.D. student at the School of Computer Science, Wuhan University, Wuhan, China. Her research
interests focus on Federated Learning.
\end{IEEEbiography}

\begin{IEEEbiography}[{\includegraphics[width = 1in, height = 1.25in, clip, keepaspectratio]{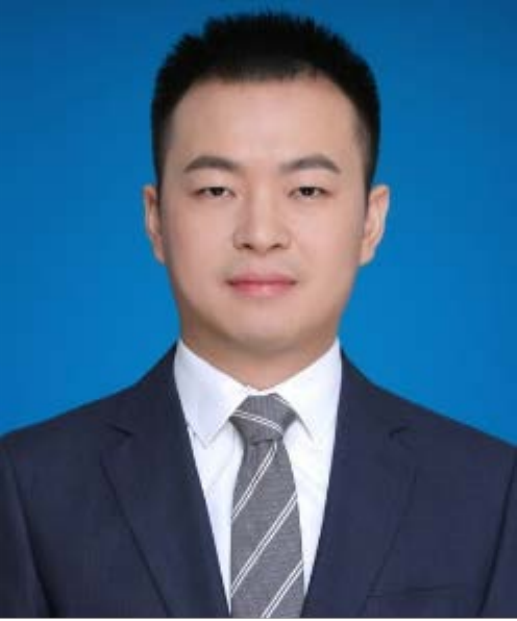}}]{Mang Ye}
received the Ph.D. degree in Computer Science from Hong Kong Baptist University, in 2019. He was a Research Scientist at Inception Institute of Artificial Intelligence. He is currently a full professor with the School of Computer Science, Wuhan University, Wuhan, China. He has published more than 100 articles in top-tier venues. He serves as Area Chair of CVPR 2024, ECCV 2024 and NeurIPS 2024. He serves as Associate Editor for IEEE Transactions on Information Forensics and Security (TIFS) and IEEE Transactions on Image Processing (TIP). His research interests focus on computer vision, pattern recognition, federated learning.
\end{IEEEbiography}

\begin{IEEEbiography}[{\includegraphics[width = 1in, height = 1.25in, clip, keepaspectratio]{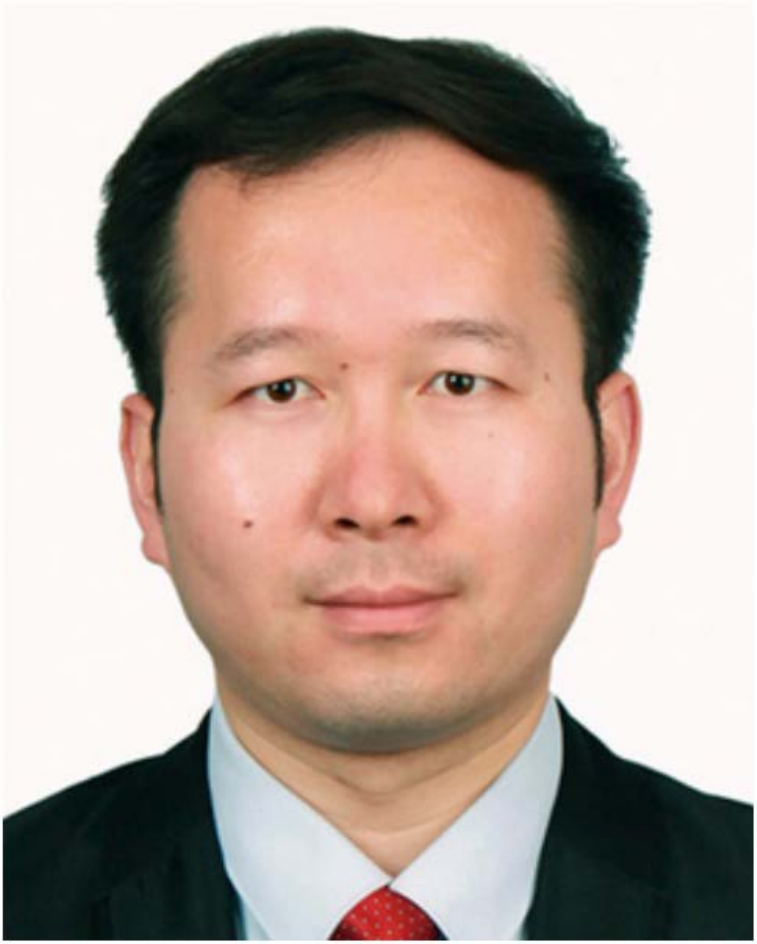}}]{Bo Du }
received the Ph.D. degree in photogrammetry and remote sensing from the State Key Laboratory of Information Engineering in Surveying, Mapping and Remote Sensing, Wuhan University, Wuhan, China, in 2010.,He is a Professor with the School of Computer Science, Wuhan University. He has published over 200 research articles. His major research interests include pattern recognition, hyperspectral image processing, machine learning, and signal processing.,Dr. Du was a recipient of the Distinguished Paper Award from IJCAI 2018, the Best Paper Award of the IEEE Whispers 2018, the Champion Award of the IEEE Data Fusion Contest 2018, the Best Reviewer Award from the IEEE GRSS for his service to the IEEE Journal of Selected Topics in Earth Observations and Applied Remote Sensing (JSTARS) in 2011, and the ACM rising star awards for his academic progress in 2015. He was the Session Chair of the International Geoscience and Remote Sensing Symposium (IGARSS) 2018/2016 and the 4th IEEE GRSS Workshop on Hyperspectral Image and Signal Processing: Evolution in Remote Sensing (WHISPERS). 
\end{IEEEbiography}

\clearpage
\noindent\textbf{\large Supplemental Materials:}

\setcounter{equation}{0}
\setcounter{figure}{0}
\setcounter{table}{0}
\setcounter{page}{1}
\makeatletter
\renewcommand{\theequation}{R\arabic{equation}}
\renewcommand{\thefigure}{R\arabic{figure}}
\renewcommand{\thetable}{R\arabic{table}}

\definecolor{mygray}{gray}{.9}
\begin{table*}[b]\small
\setlength{\abovecaptionskip}{0cm}
\centering
\caption{\blue{Comparison of AsymHFL performance under different knowledge transfer matrix update frequency with corruption rate $\xi=1$ on private dataset. $T_f$ denotes that the knowledge transfer matrix is updated once every $T_f$ rounds of collaborative learning. $\theta_k$ represents the local model of the client $c_k$.}}
\label{tab:asymhflfre}
\begin{tabular}{P{1.2cm}||P{1.2cm}P{1.2cm}P{1.2cm}P{1.2cm}|P{1.2cm}||P{1.2cm}P{1.2cm}P{1.2cm}P{1.2cm}|P{1.2cm}}
\hline
\specialrule{.1em}{0em}{0em}
\rowcolor{mygray}
    & \multicolumn{5}{c||}{Test on clean dataset} & \multicolumn{5}{c}{Test on random corrupted dataset}\\
\hhline{*1>{\arrayrulecolor[gray]{.9}}->{\arrayrulecolor{black}}||----|-||----|-}
\rowcolor{mygray}
    \multirow{-2}{*}{$T_f$} & $\theta_0$ & $\theta_1$ & $\theta_2$ & $\theta_3$ & Avg & $\theta_0$ & $\theta_1$ & $\theta_2$ & $\theta_3$ & Avg\\
\hhline{-||----|-||----|-}
    1 & 79.07& 80.37& \textbf{71.69}& \textbf{79.47}& \textbf{77.65}& \textbf{76.27}& 74.50& \textbf{69.42}& \textbf{75.67}& \textbf{73.96}\\
    2 &	77.79& \textbf{81.13}& 71.34& 78.23& 77.12& 74.55& 76.28& 68.80& 73.82& 73.36\\
    3 & \textbf{79.12}& 80.14& 70.67& 77.26& 76.80& 75.70& \textbf{77.31}& 68.01& 73.30& 73.58\\
    4 & 76.50& 79.26& 69.71& 76.91& 75.59& 73.92& 75.79& 67.85& 72.91& 72.62\\
    5 & 78.50& 79.71& 69.65& 76.21& 76.02& 74.41& 76.42& 66.57& 73.03& 72.61\\
\hline
\specialrule{.1em}{0em}{0em}
\end{tabular}
  \vspace{-2mm}
\end{table*}

\subsection*{A. Matrix Update Frequency Analysis}
\blue{We conduct comprehensive experiments and analysis on different knowledge transfer matrix update frequencies. Specifically, we set the knowledge transfer matrix to be updated once every $T_f$ rounds, $T_f\in\{1,2,3,4,5\}$. The experimental results (Table~\ref{tab:asymhflfre}) show that the more frequently the knowledge transfer matrix $M$ is updated, the more accurate the model obtained by the AsymHFL algorithm. While less frequent updates could reduce computational overhead, they also result in suboptimal performance. Specifically, when $T_f$ increases from 1 to 5, the average results of the model tested on the clean dataset and on the randomly damaged dataset dropped by 1.63\% and 1.35\% respectively. Therefore, we chose to update the knowledge transfer matrix $M$ in each round of collaborative learning to obtain the best algorithm performance.}

\begin{table*}[b]\small
\setlength{\abovecaptionskip}{0cm}
\centering
\caption{Compare with the SOTA methods with corruption rate $\xi=0.5$ on non-IID private dataset, $\theta_k$ represents the local model of the client $c_k$.}
\label{tab:sotanoniid}
\begin{tabular}{P{2.3cm}||P{1.1cm}P{1.1cm}P{1.1cm}P{1.1cm}|P{1.1cm}||P{1.1cm}P{1.1cm}P{1.1cm}P{1.1cm}|P{1.1cm}}
\hline
\specialrule{.1em}{0em}{0em}
\rowcolor{mygray}
    & \multicolumn{5}{c||}{Test on clean dataset} & \multicolumn{5}{c}{Test on random corrupted dataset}\\
\hhline{*1>{\arrayrulecolor[gray]{.9}}->{\arrayrulecolor{black}}||----|-||----|-}
\rowcolor{mygray}
    \multirow{-2}{*}{Model} & $\theta_0$ & $\theta_1$ & $\theta_2$ & $\theta_3$ & Avg & $\theta_0$ & $\theta_1$ & $\theta_2$ & $\theta_3$ & Avg\\
\hhline{-||----|-||----|-}
    Baseline & 63.13& 72.17& 66.30& 59.69& 65.32& 59.51& 66.48& 60.64& 55.40& 60.51\\
    FedMD\cite{nips2019fedmd} & 66.24& 76.05& 70.65& 65.18& 69.53& 62.63& 70.70& 65.66& 60.90& 64.97\\
    FedDF\cite{nips2020feddf} & 66.15& 71.11& 70.58& 64.91& 68.19& 62.62& 65.30& 66.31& 60.78& 63.75\\
    KT-pFL\cite{nips2022ktpfl} & 67.01& 77.83& 68.23& 65.72& 69.70& 63.39& 71.98& 63.52& 61.19& 65.02\\
    RHFL\cite{cvpr2022rhfl} & 65.08& 76.08& 67.81& 61.74& 67.68& 59.85& 70.44& 62.38& 57.69& 62.59\\
    FCCL\cite{cvpr2022fccl} & 67.15& 75.24& 71.09& 64.81& 69.57& 63.41& 70.11& 66.02& 60.22& 64.94\\
    FedProto\cite{aaai2022fedproto} & 66.66& \textbf{78.39}& 68.95& 65.37& 69.84& 63.18& 72.36& 64.21& 61.25& 65.25\\
\hhline{-||----|-||----|-}
    AugHFL~\cite{iccv2023aughfl} & 66.63& 75.89& \textbf{73.11}& 63.77& 69.85& 63.81& 71.71& 69.72& 60.49& 66.43\\
    RAHFL &  \textbf{68.32}& 76.21& 73.03& \textbf{66.47}& \textbf{71.01}& \textbf{65.19}& \textbf{72.44}& \textbf{69.95}& \textbf{62.50}& \textbf{67.52}\\
\hline
\specialrule{.1em}{0em}{0em}
\end{tabular}
  \vspace{-2mm}
\end{table*}

\subsection*{B. Data Heterogeneity Analysis}
\begin{figure}[h]
\centering
   \includegraphics[width=8.8cm]{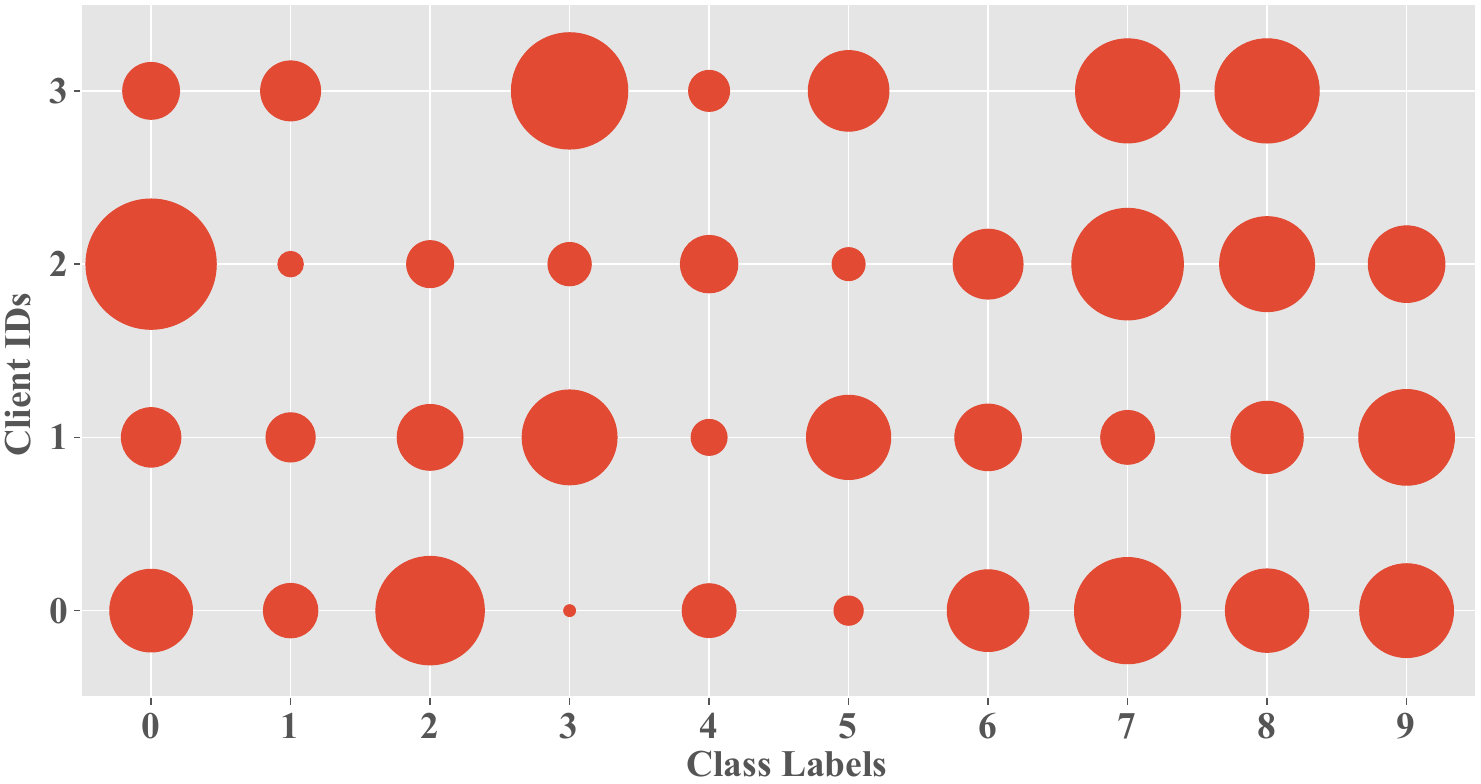}
   \vspace{-1mm}
   \caption{\small{Illustration of the number of samples per class in each client when the Dirichlet distribution beta value is 1.0. The size of each point represents the quantity of samples for each class in the respective client.
   }}
   \vspace{-4mm}
\label{fig:noniid}
\end{figure}

We conduct additional experiments to evaluate the performance of RAHFL in data heterogeneous FL scenarios, where the local data of clients are not Independent and Identically Distributed (non-IID). In the non-IID setting, the label distribution of each client follows a Dirichlet distribution. The samples for each client are sampled from the entire training dataset according to the label distribution following the Dirichlet distribution. The hyperparameter $\beta$ of the Dirichlet distribution controls the degree of data heterogeneity, with smaller values of $\beta$ indicating higher heterogeneity. For our experiment, We set $\beta$ to $1.0$ to explore the effectiveness of our method in data heterogeneous scenarios. Figure~\ref{fig:noniid} shows the number of samples per class in each client under this setting.
\blue{Table~\ref{tab:sotanoniid} shows the comparison results with SOTA methods under a corruption rate of $\xi=0.5$.} 

The results demonstrate that both AugHFL and RAHFL achieve significant improvements over existing methods in tests on both clean and corrupted datasets, especially RAHFL. Specifically, when tested on the clean dataset, RAHFL achieves an average accuracy of 71.01\%, which is 5.69\% higher than the baseline and more than 1.16\% higher than other methods. When tested on a randomly corrupted dataset, RAHFL maintains strong performance, with an average accuracy of 67.52\%, 7.01\% higher than the baseline and over 1.09\% higher than other methods. These results highlight the effectiveness of RAHFL in adapting to non-IID data distributions, thereby enhancing its practicality in real-world FL scenarios. We attribute this adaptability to AsymHFL, which can customize the aggregation of external knowledge based on local data characteristics. In AsymHFL, we use global generic test data to measure client model performance, encouraging clients to learn from those with uniform data distribution, high data quality, and strong model performance. By allowing local models to focus on relevant knowledge specific to their local data, RAHFL mitigates the negative impact of data heterogeneity on FL performance. In addition, the DCL module in RAHFL promotes robust representation learning, further improving the generalization and adaptability of the model under various data conditions. These modules together improve the performance and robustness of RAHFL in data heterogeneous FL environments.

\begin{figure*}[t]
\centering{
    \subcaptionbox{$\xi=0$; test on clean dataset}{\includegraphics[width=0.33\linewidth]{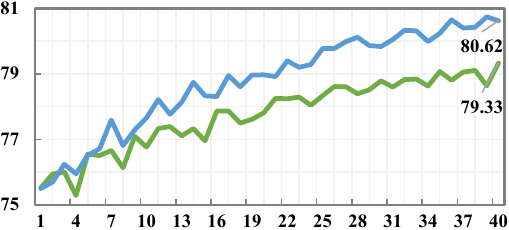}}
    \subcaptionbox{$\xi=0.5$; test on clean dataset}
    {\includegraphics[width=0.33\linewidth]{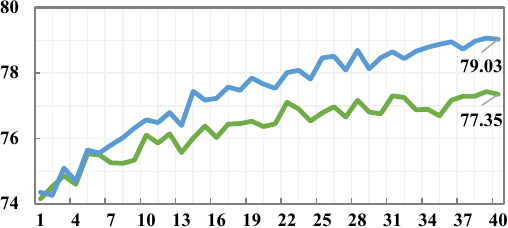}}
    \subcaptionbox{$\xi=1$; test on clean dataset}{\includegraphics[width=0.33\linewidth]{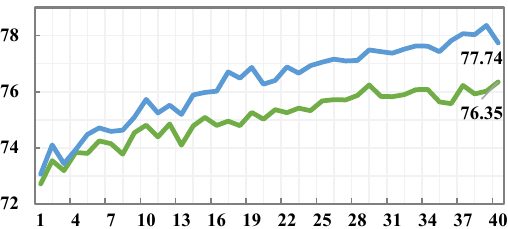}}
    \subcaptionbox{$\xi=0$; test on corrupted dataset}{\includegraphics[width=0.33\linewidth]{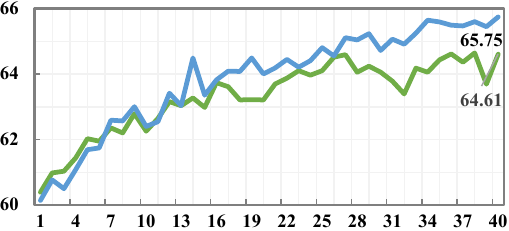}}
    \subcaptionbox{$\xi=0.5$; test on corrupted dataset}{\includegraphics[width=0.33\linewidth]{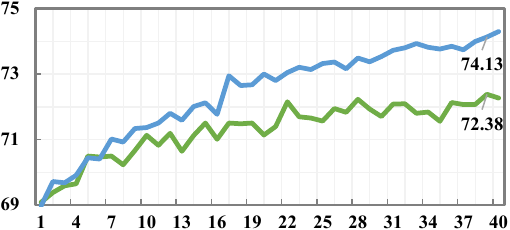}}
    \subcaptionbox{$\xi=1$; test on corrupted dataset}{\includegraphics[width=0.33\linewidth]{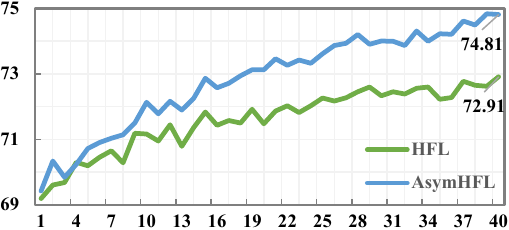}}
    \vspace{-1mm}
    \caption{Comparison of accuracy between HFL and AsymHFL during the FL process with 20 clients. The x-axis represents the communication epochs, and the y-axis denotes the test accuracy. The performance at the final communication epoch is highlighted for each method.}
    \label{fig:20lients}}
  \vspace{-4mm}
\end{figure*}

\subsection*{C. Performance Evaluation with Numerous Clients}
To demonstrate the effectiveness of our approach in scenarios involving numerous clients, we conduct an extensive performance evaluation using 20 clients to speculate on its scalability and robustness in large-scale FL environments. Our experimental setup includes 20 clients with heterogeneous local models participating in the FL process, where ResNet10, ResNet12, ShuffleNet, and MobileNetv2 are assigned to five clients respectively. Since the Aug and DCL modules act on the local update phase and are unaffected by the number of participating clients, we focus solely on evaluating the performance of the AsymHFL module in scenarios with numerous clients.
The results are shown in Figure~\ref{fig:20lients}, which compares the accuracy of HFL and AsymHFL throughout the FL process under various settings. The x-axis in the figure represents the communication time and the y-axis represents the test accuracy. We show the average test accuracy of all clients after each collaborative learning round.
The results show that even with numerous clients, AsymHFL consistently outperforms the traditional HFL method throughout the communication cycle and achieves higher average test accuracy in almost all rounds. In standard HFL, clients perform symmetric knowledge exchange, which may result in the propagation of low-quality information from poorly performing clients, reducing the overall learning efficiency. This will slow down convergence and reduce the accuracy of the final model. However, AsymHFL enables clients to selectively and asymmetrically learn from other clients, effectively filtering out corrupted information and promoting higher quality and more efficient knowledge transfer. As a result, compared with general HFL, AsymHFL exhibits superior effectiveness and stability throughout the training process, achieving higher accuracy in fewer cycles. This performance evaluation in scenarios with numerous clients highlights the practical effectiveness of our approach in large-scale FL environments.
\end{document}